\documentclass{article}

\usepackage[top=1in, bottom=1in, left=1in, right=1in]{geometry}
\usepackage[utf8]{inputenc} 
\usepackage[T1]{fontenc}    
\usepackage{hyperref}       
\usepackage{url}            
\usepackage{booktabs}       
\usepackage{amsfonts}       
\usepackage{nicefrac}       
\usepackage{xcolor}         
\usepackage{bbm}
\usepackage{amsmath}
\usepackage{mathtools}
\usepackage{graphicx}
\usepackage{subcaption}
\usepackage{float}
\usepackage{amssymb}
\usepackage{gensymb}
\usepackage{amsthm}
\usepackage{multirow}
\usepackage{adjustbox}
\usepackage[title]{appendix}
\usepackage[ruled]{algorithm2e}

\usepackage{tikz}
\usepackage{tikz-cd}
\usepackage{pgfplots}
\usepackage{placeins}
\usepackage{natbib}
\usepackage{setspace}
\usetikzlibrary{patterns, shapes.geometric, positioning, arrows, arrows.meta, shadows, matrix, fit, backgrounds}
\pgfplotsset{compat=1.14}

\usepackage{thmtools,thm-restate}

\DeclareMathOperator{\ind}{\mathbbm{1}}
\DeclareMathOperator{\Pa}{Pa}
\DeclareMathOperator{\pa}{pa}

\DeclareMathOperator{\rank}{Rank}
\DeclareMathOperator{\hit}{HR@}
\DeclareMathOperator{\agg}{agg}
\DeclareMathOperator{\argmax}{arg\,max}
\DeclareMathOperator{\eff}{eff}
\DeclareMathOperator{\grid}{Grid}
\DeclareMathOperator{\toolpark}{TP}
\DeclareMathOperator{\util}{UT}
\DeclareMathOperator{\batcont}{BC}
\DeclareMathOperator{\batusage}{BU}
\DeclareMathOperator{\batsoc}{SOC}
\DeclareMathOperator{\cooling}{CL}
\DeclareMathOperator{\daytime}{DT}

\newcommand{\abbreviation}[2]{\textit{#2 (#1)}}

\newcommand{\defeq}{\coloneqq} 
\newcommand{\Rn}{\mathbb{R}} 
\newcommand{\Nn}{\mathbb{N}} 
\newcommand{\Zn}{\mathbb{Z}} 
\usepackage[font={footnotesize}]{caption}

\newcommand{\ABSTRACT}[1]{\begin{abstract}#1\end{abstract}}
\newcommand{\KEYWORDS}[1]{\textbf{Keywords:} #1}

\title{Causal explanations of outliers in systems with lagged time-dependencies}

\author{
  Philipp Alexander Schwarz\footnotemark[1] \\
  AMS-Osram \\
  University of Hamburg \\
  \and
  Johannes Oberpriller \\
  AMS-Osram \\
  \and
  Sven Klaassen \\
  University of Hamburg \\
  Economic AI\\
}
\date{}

\begin{document}
\maketitle
\ABSTRACT{%
Root-cause analysis in controlled time dependent systems poses a major challenge in applications.
Especially energy systems are difficult to handle as they
exhibit instantaneous as well as delayed effects and if equipped with storage, do have a memory.
In this paper we adapt the causal root-cause analysis method
of \cite{Budhathoki22a} to general time-dependent systems,
as it can be regarded as a strictly causal definition of the term ``root-cause''.
Particularly, we discuss two truncation approaches to handle the infinite dependency graphs
present in time-dependent systems.
While one leaves the causal mechanisms intact,
the other approximates the mechanisms at the start nodes.
The effectiveness of the different approaches is benchmarked 
using a challenging data generation process inspired
by a problem in factory energy management: the avoidance of peaks in the power consumption.
We show that given enough lags our extension
is able to localize the root-causes in the feature and time domain.
Further the effect of mechanism approximation is discussed.
}%


\doublespacing
\section{Introduction}
Explaining the behavior of dynamical systems poses a major challenge in contemporary research.
In recent years data-driven methods have fostered the analysis and optimization of complex systems
in various research branches, like earth sciences \citep{Runge2019, Runge2019a},
energy management \citep{Guo2024} and epidemiology \citep{Borges2025}.
Particularly, the problem of identifying causes of anomalies in such systems has gained major attraction.
While most of this attention is coming from cloud-computing \citep{Wang2023, Liu2021, Lin2024, Ikram2022}
the methodology also finds application areas like healthcare \cite{Strobl2024}.
In general, \abbreviation{RCA}{Root Cause Analysis} tries to locate the sources of 
outlier events with the goal to use the resulting explanations to take actions
that steer the system in question back to a normal state, or to avoid similar anomalies in the future.
We follow the view of \cite{Budhathoki22a} that in these cases explanations should be derived from
causal relations affecting the target variable.
That is, the applied methodology should be based on counterfactual reasoning.
For a review of the three main methodology groups, namely correlation-based, causal and hybrid approaches, we refer to \cite{Dawoud2025}.

Further we argue that complex systems may exhibit lagged time-dependencies 
that make RCA especially challenging:
\begin{enumerate}
  \item The system may contain variables that have an instantaneous and
    others that have a delayed effect on the target,
    a property that prohibits simplification of time-series data through coarse aggregations
    \citep{Runge2018}.

  \item The systems target variable may be controlled or self-regulating
    which implies that anomalies are seldom rooted in a single cause.
    That is, outlier behavior of a subsystem may only propagate to the systems target
    if the control mechanism is also faulty. 

  \item The system may have a memory which implies potential long-ranging effects on its target.
\end{enumerate}
In order to deal with these challenges we adopt the
\abbreviation{CRCA}{causal structur-based root cause analysis} method from \cite{Budhathoki22a} 
to lagged time-dependencies.
This choice is deliberate and in contrast to existing methods that deal with time-series data.
\cite{Lin2024} and \cite{Wang2023}, for example, rely on weaker forms of causality, that is Granger causality,
but employ advanced machine learning techniques to estimate the causal structure.
Our rational is that CRCA can be viewed as an attempt to formalize the notion of root-causes
in the framework of structural causal models.
In this sense the present study can be seen as an evaluation of the basic definition of root causes 
in time dependent domains.

As an example and benchmark serves a \abbreviation{DGP}{data-generating process} modeling
the electrical system of a hypothetical manufacturing plant.
It consists of energy consumers (e.g. tool parks), stores (e.g. battery) and producers,
and is connected to the public grid.
The power draw from the public grid is regulated not to exceed a certain limit,
such that consumption peaks exceeding this limit can be regarded as anomalies.
Note that, a machine may increase its power consumption only with a delay of several minutes 
after the start of production, whereas the loading and unloading of a battery storage
affects the power draw almost immediately.
A behavior that is represented by the DGP.
In summary the studied DGP exhibits the challenges mentioned before.

\section{Methodology}\label{sec:methodology}
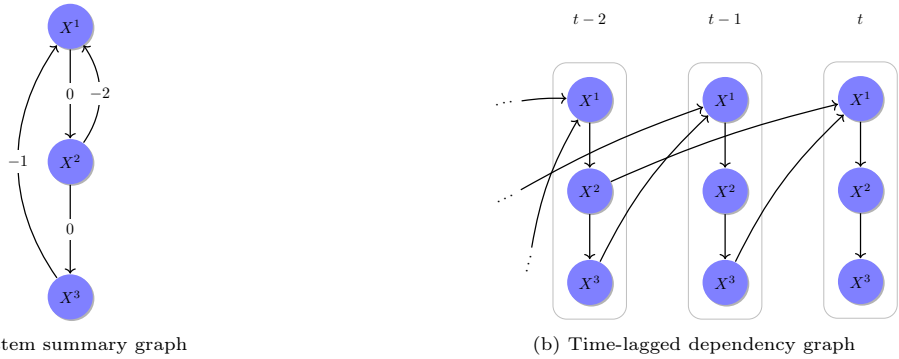
\begin{figure}[H]
  \centering
  \begin{subfigure}{0.5\linewidth}
    \scalebox{0.6}{\begin{tikzpicture}
\tikzset{
  mea/.style={fill=blue!50,general shadow={fill=gray!60,shadow xshift=1pt,shadow yshift=-1pt}},
  cont/.style={fill=red!60,general shadow={fill=gray!60,shadow xshift=1pt,shadow yshift=-1pt}},
  proc/.style={fill=yellow!20},
  inv/.style={fill=white}
}

\begin{scope}[
  thick,
  local bounding box=graph,
  every node/.style={circle},
  label/.style={rectangle,fill=white},
  control/.style={cont,circle,minimum size=1cm},
  measurement/.style={mea,circle,minimum size=1cm},
]
  \node[measurement] (X1) at (0, 6) {\(X^1\)};
  \node[measurement] (X2) at (0, 3) {\(X^2\)};
  \node[measurement] (X3) at (0, 0) {\(X^3\)};

  \path [->] (X1) edge node [label] {\(0\)} (X2);
  \path [->] (X2) edge node [label] {\(0\)} (X3);

  \path [->,bend left=35] (X3) edge node [label] {\(-1\)} (X1);
  \path [->,bend right=35] (X2) edge node [label] {\(-2\)} (X1);
\end{scope}
\end{tikzpicture}}
    \centering
    \caption{System summary graph}
    \label{fig:dep_graph_summary}
  \end{subfigure}%
  ~
  \begin{subfigure}{0.5\linewidth}
    \scalebox{0.6}{\begin{tikzpicture}

\tikzset{
  mea/.style={fill=blue!50,general shadow={fill=gray!60,shadow xshift=1pt,shadow yshift=-1pt}},
  cont/.style={fill=red!60,general shadow={fill=gray!60,shadow xshift=1pt,shadow yshift=-1pt}},
  proc/.style={fill=white!20,draw=lightgray},
  inv/.style={fill=white}
}

\begin{scope}[
  thick,
  local bounding box=graph,
  every node/.style={circle},
  label/.style={rectangle},
  control/.style={cont,circle,minimum size=1cm},
  measurement/.style={mea,circle,minimum size=1cm},
]
  \node[label] (Pml) at (13, 0) {\(t\)};
  \node[measurement,below=of Pml] (X1l0) {\(X^1\)};
  \node[measurement,below=of X1l0] (X2l0) {\(X^2\)};
  \node[measurement,below=of X2l0] (X3l0) {\(X^3\)};
  \path [->] (X1l0) edge node {} (X2l0);
  \path [->] (X2l0) edge node {} (X3l0);

  \node[label] (Pml1) at (10, 0) {\(t-1\)};
  \node[measurement,below=of Pml1] (X1l1) {\(X^1\)};
  \node[measurement,below=of X1l1] (X2l1) {\(X^2\)};
  \node[measurement,below=of X2l1] (X3l1) {\(X^3\)};
  \path [->] (X1l1) edge node {} (X2l1);
  \path [->] (X2l1) edge node {} (X3l1);

  \node[label] (Pml2) at (7, 0) {\(t-2\)};
  \node[measurement,below=of Pml2] (X1l2) {\(X^1\)};
  \node[measurement,below=of X1l2] (X2l2) {\(X^2\)};
  \node[measurement,below=of X2l2] (X3l2) {\(X^3\)};
  \path [->] (X1l2) edge node {} (X2l2);
  \path [->] (X2l2) edge node {} (X3l2);
\end{scope}

\begin{scope}[
  thick,
  every node/.style={fill=white,circle,text opacity=1,fill opacity=0},
  every edge/.style={fill=white,text opacity=0,fill opacity=0,very thick},
]
  \node[black,left=of X2l2,rotate=25] (X2l3) {\(\cdots\)};
  \node[black,below=of X2l3,rotate=75] (X3l3) {\(\dots\)};
  \node[black,left=of X1l2,rotate=12] (X2l4) {\(\cdots\)};
\end{scope}

\begin{scope}[
  on background layer,
  thick,
  process/.style={proc,rectangle,rounded corners=2ex},
]
  \node[process, fit=(X1l2) (X2l2) (X3l2), inner sep=3mm] (Pm) {};
  \node[process, fit=(X1l1) (X2l1) (X3l1), inner sep=3mm] (Pm) {};
  \node[process, fit=(X1l0) (X2l0) (X3l0), inner sep=3mm] (Pm) {};
\end{scope}

\begin{scope}[
  thick,
]
  \path[->, bend left=10] (X3l1) edge (X1l0);
  \path[->, bend left=5] (X2l2) edge (X1l0);

  \path[->, bend left=10] (X3l2) edge (X1l1);
  \path[->, bend left=5] (X2l3) edge (X1l1);

  \path[->, bend left=10] (X3l3) edge (X1l2);
  \path[->, bend left=5] (X2l4) edge (X1l2);
\end{scope}
\end{tikzpicture}}
    \centering
    \caption{Time-lagged dependency graph}
    \label{fig:dep_graph_timelagged}
  \end{subfigure}
  \caption[Dependencies of an exemplary system]{
    Dependencies of an exemplary system with
    \(X^1_t \defeq f_1(X^2_{t-2}, X^3_{t-1}\,|\, N^1)\), 
    \(X^2_t \defeq f_2(X^1_t\,|\,N^2)\)
    and 
    \(X^3_t \defeq f_3(X^2_t\,|\,N^3)\).
  }
  \label{fig:dep_graph}
\end{figure}

In this section we introduce the basic causal setting and adapt the CRCA method to it.
Particularly, regarding causality, we rely on the framework of \cite{Pearl2022},
and make use of \abbreviation{SCMs}{causal structural models} as defined in \citep{Peters2017}.
Note that time series have been studied before in this context (e.g. \citep{Runge2018}).
We use the notation
\(\Zn_{\leq t} \defeq \left\{s \in \Zn \,\middle|\, s \leq t \right\}\)
for the set of integers lower or equal to \(t\).
Let \(X_t \defeq \left(X^1_t, \ldots, X^n_t\right)\) be a discrete-time multivariate process
that propagates in time according to the causal structural model
\begin{align}\label{eq:scm_functional}
  X^i_t \defeq f^i\left(\Pa^i_t,\, N^i_t\right)
\end{align}
where 
\(N_t^i\) denote the noise variables
and
\(\Pa^i_t \subset \{X^i_s \,|\, s \in \Zn_{\leq t}, \, i = 1, \ldots, n\} \setminus \{X^i_t\}\)
the sets of parent dependencies.
We suppose that the parent dependencies \(\Pa^i_t\) are finite, acyclic and do not change with time.
For fixed \(i\) let the noise terms \(N^i_t\) be time independent
and identical distributed, such that 
\(X^i_t = f^i\left(\Pa^i_t,\, N^i\right)\) holds with \(N^i \defeq N^i_0\).
Under these assumptions we can summarize the infinite time dependency graph 
of \(X_t\) as shown in Figure \ref{fig:dep_graph}.
Further we assume the components of the noise vector \(N \defeq \left(N^1,\, \ldots,\, N^n\right)\)
to be jointly independent in accordance to the definition the SCM.
The lower case notation (e.g. \(n^i_t\) and \(\pa^i_t\))
is used to denote realizations of the matching capitalized random variable,
or set of random variables (e.g. \(N^i_t\) and \(\Pa^i_t\)).

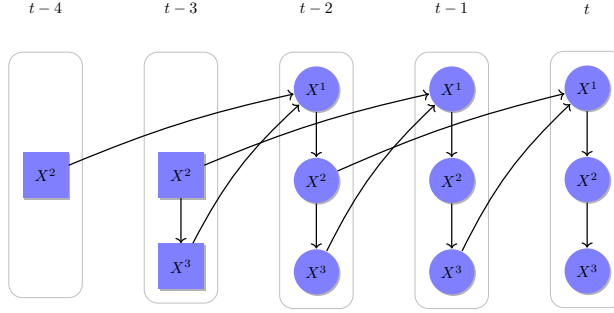
\begin{figure}
  \centering
  \scalebox{0.6}{\begin{tikzpicture}

\tikzset{
  mea/.style={fill=blue!50,general shadow={fill=gray!60,shadow xshift=1pt,shadow yshift=-1pt}},
  cont/.style={fill=red!60,general shadow={fill=gray!60,shadow xshift=1pt,shadow yshift=-1pt}},
  proc/.style={fill=white!20,draw=lightgray},
  inv/.style={fill=white}
}

\begin{scope}[
  thick,
  local bounding box=graph,
  every node/.style={circle},
  label/.style={rectangle},
  conditioned/.style={mea,rectangle,minimum size=1cm},
  measurement/.style={mea,circle,minimum size=1cm},
]
  \node[label] (Pml) at (13, 0) {\(t\)};
  \node[measurement,below=of Pml] (X1l0) {\(X^1\)};
  \node[measurement,below=of X1l0] (X2l0) {\(X^2\)};
  \node[measurement,below=of X2l0] (X3l0) {\(X^3\)};
  \path [->] (X1l0) edge node {} (X2l0);
  \path [->] (X2l0) edge node {} (X3l0);

  \node[label] (Pml1) at (10, 0) {\(t-1\)};
  \node[measurement,below=of Pml1] (X1l1) {\(X^1\)};
  \node[measurement,below=of X1l1] (X2l1) {\(X^2\)};
  \node[measurement,below=of X2l1] (X3l1) {\(X^3\)};
  \path [->] (X1l1) edge node {} (X2l1);
  \path [->] (X2l1) edge node {} (X3l1);

  \node[label] (Pml2) at (7, 0) {\(t-2\)};
  \node[measurement,below=of Pml2] (X1l2) {\(X^1\)};
  \node[measurement,below=of X1l2] (X2l2) {\(X^2\)};
  \node[measurement,below=of X2l2] (X3l2) {\(X^3\)};
  \path [->] (X1l2) edge node {} (X2l2);
  \path [->] (X2l2) edge node {} (X3l2);

  \node[label] (Pml3) at (4, 0) {\(t-3\)};
  \node[white,inv,below=of Pml3] (X1l3) {\(X^1\)};
  \node[conditioned,below=of X1l3] (X2l3) {\(X^2\)};
  \node[conditioned,below=of X2l3] (X3l3) {\(X^3\)};
  \path[->, bend left=0] (X2l3) edge (X3l3);

  \node[label] (Pml4) at (1, 0) {\(t-4\)};
  \node[white,inv,below=of Pml4] (X1l4) {\(X^1\)};
  \node[conditioned,below=of X1l4] (X2l4) {\(X^2\)};
  \node[white,inv,below=of X2l4] (X3l4) {\(X^3\)};

\end{scope}

\begin{scope}[
  thick,
  every node/.style={fill=white,circle,text opacity=1,fill opacity=0},
  every edge/.style={fill=white,text opacity=0,fill opacity=0,very thick},
]
  %

\end{scope}

\begin{scope}[
  on background layer,
  thick,
  process/.style={proc,rectangle,rounded corners=2ex},
]
  \node[process, fit=(X1l2) (X2l2) (X3l2), inner sep=3mm] (Pm) {};
  \node[process, fit=(X1l1) (X2l1) (X3l1), inner sep=3mm] (Pm) {};
  \node[process, fit=(X1l0) (X2l0) (X3l0), inner sep=3mm] (Pm) {};

  \node[process, fit=(X1l3) (X2l3) (X3l3), inner sep=3mm] (Pm) {};
  \node[process, fit=(X1l4) (X2l4) (X3l4), inner sep=3mm] (Pm) {};
\end{scope}

\begin{scope}[
  thick,
]
  
  \path[->, bend left=10] (X3l1) edge (X1l0);
  \path[->, bend left=5] (X2l2) edge (X1l0);

  \path[->, bend left=10] (X3l2) edge (X1l1);
  \path[->, bend left=5] (X2l3) edge (X1l1);

  \path[->, bend left=10] (X3l3) edge (X1l2);
  \path[->, bend left=5] (X2l4) edge (X1l2);




  %




\end{scope}


\end{tikzpicture}}
  \caption[Unfolding and truncation of example graph shown in Figure \ref{fig:dep_graph_summary} with \(L=2\)]{
    Unfolding and truncation of example graph shown in Figure \ref{fig:dep_graph_summary} with \(L=2\):
    The rectangular nodes represent the dangling parents
    \(\Pa_{[t-2,t]}(3) = \{X^2_{t-4},\, X^2_{t-3},\, X^3_{t-3}\}\)
    on which we condition in the truncation model
    whereas the noise terms \(N_{[t-L,\,t]} = \{(N_{t-2}, N_{t-1}, N_t)\}\)
    of the circular nodes are assessed in the attribution analysis.
    In the non-truncated case the mechanisms and noise distributions
    of the dangling nodes are adapted the remaining dependencies
    (e.g. \(X^2_{t-3} \to X^3_{t-3}\) remains while  \(X^2_{t-3}\) has no more parents)
    and then also employed for the attribution analysis.
  }
  \label{fig:example_unfold}
\end{figure}

We follow \cite{Budhathoki22a} and view \(X^n_t\) as the systems target,
for which we want to explain anomalous behavior.
That is, for which we want to localize root-causes among the individual
time series \(X^1_t,\, \ldots, \, X^n_t\).
To apply the CRCA methodology, it is necessary to limit the search to a certain  
time interval \([t-L,\, t]\) where \(L \in \Zn_{\geq 0}\)
denotes the maximum time-lag considered.
Choosing a finite maximum lag narrows the root-cause identification down to
the more immediate past excluding very long range effects. 
Especially in systems with a memory choosing \(L\) is a tradeoff between relevance, 
accuracy and computational time.
Let 
\[\Pa^i_t(l) \defeq \{X^j_s \,|\, s \in \Zn_{\leq t-l},\, j=1, \ldots, n\} \cap \Pa^i_t\]
and respectively
\[\Pa_t(l) \defeq \Pa^1_t(l) \cup \cdots \cup \Pa^n_t(l)\]
denote the parent dependencies of \(X^i_t\) and \(X_t\) with minimum time-lag \(l > 0\).
Through a recursive argument one can show that \(X_t^n\) can be written as
a function \(F_L\) of noise terms 
\(N_{[t-L,\, t]} \defeq \left(N_{t-L},\, \ldots,\, N_t\right)\)
up to lag \(L\) and 
parent dependencies 
\(\Pa_{[t-L,\, t]}(L+1) \defeq \bigcup_{l=0}^L \Pa_{t-l}(L-l+1) \)
that exceed \(L\):
\begin{align*}
  X_t^n = F_L\left(\Pa_{[t-L,\,t]}(L+1), \, N_{[t-L,\,t]}\right)
\end{align*}
This procedure can be seen as an \(L\)-times unfolding of the summary graph of the system (see Figure \ref{fig:example_unfold}),
followed by a truncation at the dangling parents
\(\Pa_{[t-L,\, t]}(L+1)\)
and puts us in a position to apply CRCA up to nodes of lag \(L\).
The basic idea of CRCA is that extreme noise realizations are at fault for an anomaly.
Thus, the question asked to locate the root-causes is:
would an anomaly still be probable if certain nodes had received
normal noise values according to their distributions instead of the observed?
More precisely we suppose that at time \(t\) a system anomaly \(x_t^n\) 
was observed with noises \(n_{[t-L, t]}\) and dangling parents \(\pa_{[t-L,\,t]}(L+1)\).
Then the attributions of nodes
\(X_{t-l}^i\) with \(l=1, \ldots, L\) and  \(i=1, \ldots, n\)
are calculated by probing the structural model
\begin{align}\label{eq:meth:lagged_trunc_scm}
  n_L \mapsto F_L\left(\pa_{[t-L,\,t]}(L+1), \; n_L\right),
\end{align}
combining the observed noise terms \(n_{[t-L,t]}\) with randomly sampled \(n_L \sim N_{[t-L,\,t]}\). 
We refer to this SCM as the \textit{truncated} lag \(L\) model.
Note that the described truncation procedure leads to an equal assessment
of the lags of \(X^i_t\) as the generating mechanism \(f^i\) is kept intact up to the max lag \(L\),
by conditioning on the observed dangling parents \(\pa_{[t-L,\,t]}(L+1)\).
This stands in contrast to the \textit{non-truncated} lag \(L\) model
for which the mechanisms \(f^j\) related to the dangling variables
\(X^j_{t-l} \in \Pa_{[t-L,\, t]}(L+1)\)
modified to fit the remaining parents
\[X^j_{t-l} = h^j_l\left(\Pa_{[t-L,\,t]}(L+1) \cap \Pa^j_{t-l}, \, \hat{N}_{t-l}^j\right) \]
approximating the dependency structure of the actual structural model.
As indicted by the notation this approximation may have an effect on the noise distribution.
Again using a recursive argument on \(h^j_l\) we obtain the corresponding SCM
\begin{align}\label{eq:meth:lagged_nontrunc_scm}
  (n_D,\, n_L) \mapsto F_L\left(H_L(n_D), \; n_L\right)
\end{align}
where \(H_L\) maps the noises \(\hat{N}_{[t-l,t]}(L+1)\) of the dangling parents
under the changed functional model to their values:
\[\Pa_{[t-L, t]}(L+1) = H_L\left(\hat{N}_{[t-L, t]}(L+1)\right)\]
This procedure approximates the causal mechanisms 
and noise distributions of the initial dependencies in the unfolded graph.
In the attribution calculation the noise terms of the dangling parents
are then considered in addition to the noises up to lag \(L\).
With this the non-truncated lag \(L\) model stands in complexity and 
in the number of attributable nodes between the truncated lag \(L\) and 
the truncated lag \(L_D\) model where \(L_D\) is the maximum lag among
the variables \(\Pa_{[t-L,\,t]}(L+1)\).
We investigate the performance of both variations in the application example.

For completeness, we recapture definition of the CRCA attributions introduced in \cite{Budhathoki22a}
and apply it to the above setting.
Let \(g: \mathcal{X} \to \Rn\) be an anomaly score, like the z-score,
and let \(E_t\) be the outlier event \(E_t \defeq \{g(X_t^n) \geq g(x^n_t)\}\)
then the corresponding \abbreviation{IT-Score}{information theoretic calibration} 
of \(g\) is defined as \(S(x_t^n) \defeq -\log \Pr(E_t)\).
Further, let \(\mathcal{U}\) denote set of tuples \((i,\,l)\) indexing the nodes \(X_{t-l}^i\) and 
their corresponding noise values \(N^i_{t-l}\).
Note that for the truncated lag \(L\) model we have
\(\mathcal{U} = \{1,\, \ldots,\, n\} \times \{0,\, 1,\, \ldots,\, L\}\).
In case of the non-truncated model \(\mathcal{U}\) additionally includes indices for the dangling nodes.
For a subset \(\mathcal{I} \subset \mathcal{U}\) we define
\[
  q_t(\mathcal{I}) \defeq 
  \Pr\left(E_t \,\middle|\, N^i_{t-l} = n^i_{t-l}\;\, \forall \, (i,\, l) \in \mathcal{U} \setminus \mathcal{I}\right)
\]
to be the probability of the outlier event \(E_t\) when 
randomizing the noises of the nodes \((i,\, l) \in \mathcal{I}\) according to their
distributions \(N_{t-l}^i\) and keeping the other nodes fixed to the observed values \(n^i_{t-l}\).
To evaluate these probabilities the structural models
(\ref{eq:meth:lagged_trunc_scm}) and (\ref{eq:meth:lagged_nontrunc_scm})
are used.
Further, for a node \(u \in \mathcal{U}\) let 
\[ 
  C_t\left(u \,\middle|\, \mathcal{I} \right) 
  \defeq \log\frac{
    q_t\left(\mathcal{I}\right)
  }{
    q_t\left(\mathcal{I} \cup \{u\}\right)
  }
\]
be the contribution of \(u\) under randomization of
\(\mathcal{I} \subset \mathcal{U}\setminus\{u\}\).
Then the attribution of node \(u\) to the anomaly \(x_t^n\) is then defined as
\[ 
  \phi_t(u) \defeq
  \frac{1}{|\mathcal{U}|}
  \sum_{\mathcal{I} \subset \mathcal{U}\setminus\{u\}} 
  \binom{|\mathcal{U}|-1}{|\mathcal{I}|}^{-1} C_t\left(u \,|\, \mathcal{I}\right)
\]
using Shapley values \citep{Shapley1952} to factor out the dependence of \(C_t\) on 
specific index sets.
The IT-Score decomposes into the attributions as follows:
\begin{align}\label{eq:decomposition}
  S(x_t^n) = \sum_{u \in \mathcal{U}} \phi_t(u)
\end{align}
With this method we do not only obtain the nodes at fault for an anomaly but 
also information that aids the localization in time.

To apply the method it is required that the noise terms \(n_t^i\) can be calculated
from the observed values \(x_t^i\) and \(\pa_t^i\) through inversion of \(x_t^i = f^i(\pa_t^i,\, n_t^i)\).
Further when estimating \(f^i\) (or the distribution of \(N_t^i\)) the time-independence
assumptions regarding the noise distributions ensure that
the computational and memory requirements are not dependent on \(L\).
This also applies to the memory required for the sampling of noise terms to estimate \(q_t\).
However, the \(L\)-times unfolding increases the computational complexity 
to calculate \(\phi_i\) from \(n!\) to \((Ln)!\).
We adapted the existing implementation in \cite{Sharma2020, Bloebaum2024} with minimal changes,
and thus do not make use of the mentioned independence assumptions.
In particular, we estimate a functional causal model for each node and time lag.

\section{Empirical Setting}
In this section we describe the experimental setup to evaluate the time-lagged adaption of CRCA.
Part of the contribution is a data-generating process that allows to inject defined root causes in the system.
This allows to calculate the system propagation with and without injected root causes,
and thus to evaluate CRCA.

\FloatBarrier
\subsection{DGP}
The data generating process models the energy consumption of a hypothetical manufacturing plant.
The plant is connected to an external \(\grid\) that serves as an unlimited power source.
Further it consists of two tool parks \textit{\(\toolpark_a\)} and \textit{\(\toolpark_b\) }
connected to the grid injection point.
Each tool park is composed of several machines that draw power 
from the tool park connection point.
That is, when switched on a machine consumes power according to its active profile.
If production is finished the machine turns back to an idle state
drawing power according to the idle profile.
The active profile models a ramp-up phase followed by a phase of high power draw
and finally ends with a ramp-down.
The machine specifics vary only between machines of different tool parks.
The wait times between the arrival of work items are modeled to be exponentially distributed.
Thereby the utilization level of the tool parks \(\util_a\) and \(\util_b\)
tracks the share of active machines.
Moreover, in our example plant it is required to keep the production environment at a certain temperature level.
Thus, we model the cooling power \(\cooling\) of the fab to be dependent on the outside temperature \(T\).
The cooling equipment is switched on if the average temperature of the last
three minutes was over the threshold of \(19 {\degree C}\)
and turned off if it drops below \(18 {\degree C}\).
The resulting power draw from the grid junction point is time-delayed by five minutes
and proportional to the temperature.
The temperature \(T\) is dependent on the current time of the day \(\daytime\)
as real weather data is used.
The overall consumption is regulated using an on-site battery system.
Again we use a two-point battery controller \(\batcont\) that
signals the battery to remain idle, load or unload,
depending on the average draw from the grid over the last two minutes.
That is, if the average power draw is above \(1400 kW\) the battery immediately starts
to unload and continues to do so until \(1100 kW\) are reached.
Thereby the controller maintains a state of charge \(\batsoc\)
between \(70\%\) and \(90\%\).
Thus, if \(\batsoc\) drops below \(70\%\) the battery is loaded
independently of the average power draw from the grid.
To recapitulate, the overall grid consumption is defined as the sum of the tool park,
the cooling system, and battery consumption in addition to a normal distributed noise
to include unobserved consumers or producers.
Variations in the machine, cooling system and battery consumption are also 
incorporated using an additive noise.
In this setting we regard peaks in the grid consumption that exceed \(1500 kW\) as anomalies
if they have a width of at least two time steps measured from both turning points.
The described DGP produces time series data that is aggregated on a minute level 
for each of the described features.
Further it allows injecting anomalies in each major component as discussed in Section \ref{sec:injections}.

The summary graph of the system and exemplary output is shown in Figure \ref{fig:dgp_summary_graph}
and Figure \ref{fig:dgp_data}.
\begin{figure}
  \centering
  \scalebox{0.6}{\begin{tikzpicture}
\tikzset{
  mea/.style={fill=blue!50,general shadow={fill=gray!60,shadow xshift=1pt,shadow yshift=-1pt}},
  cont/.style={fill=red!60,general shadow={fill=gray!60,shadow xshift=1pt,shadow yshift=-1pt}},
  proc/.style={fill=yellow!20},
  inv/.style={fill=white}
}

\begin{scope}[
  thick,
  local bounding box=graph,
  every node/.style={circle},
  label/.style={rectangle,fill=white},
  control/.style={cont,circle,minimum size=1cm},
  measurement/.style={mea,circle,minimum size=1cm},
]
  \node[measurement] (grid) at (-0.5, 0) {\(\grid\)};
  \node[measurement] (tpa) at (-6, 1.5) {\({\toolpark}_a\)};
  \node[measurement] (tpb) at (-3, 2) {\({\toolpark}_b\)};
  \node[measurement] (batcont) at (-0.5, 3.5) {\(\batcont\)};
  \node[measurement] (batusage) at (3, 4.5) {\(\batusage\)};
  \node[measurement] (soc) at (0, 7) {\(\batsoc\)};

  \node[measurement] (cooling) at (5, 1) {\(\cooling\)};
  \node[measurement] (temperature) at (5, 4) {\(T\)};
  \node[measurement] (daytime) at (5, 7) {\(\daytime\)};

  \node[measurement] (tpua) at (-6, 7) {\(\util_{a}\)};
  \node[measurement] (tpub) at (-3, 7) {\(\util_{b}\)};

  \path [->] (tpa) edge node [label] {\(0\)} (grid);
  \path [->] (tpb) edge node [label] {\(0\)} (grid);
  \path [->] (grid) edge node [label] {\(-2,\,-1\)} (batcont);
  \path [->] (cooling) edge node [label] {\(0\)} (grid);
  \path [->] (batusage) edge node [label] {\(0\)} (grid);
  \path [->] (batcont) edge node [label] {\(0\)} (batusage);
  \path [->] (soc) edge node [label] {\(-1\)} (batcont);
  \path [->] (batusage) edge node [label] {\(-1\)} (soc);
  \path [->] (tpua) edge node [label] {\(-1,\,0\)} (tpa);
  \path [->] (tpub) edge node [label] {\(-1,\,0\)} (tpb);
  \path [->] (daytime) edge node [label] {\(0\)} (temperature);
  \path [->] (temperature) edge node [label] {\(-7,\,\ldots,\,0\)} (cooling);
  \path [->,] (tpua) edge[loop left] node {\(-1\)} (tpua);
  \path [->,] (tpa) edge[loop left] node {\(-1\)} (tpa);
  \path [->,] (tpub) edge[loop left] node {\(-1\)} (tpub);
  \path [->,] (tpb) edge[loop left] node {\(-1\)} (tpb);
  \path [->,] (soc) edge[loop left] node {\(-1\)} (soc);
  \path [->,] (batusage) edge[loop above] node {\(-2,\,-1\)} (batusage);
  \path [->,] (batcont) edge[loop left] node {\(-1\)} (batcont);
  \path [->,] (temperature) edge[loop right] node {\(-1\)} (temperature);
\end{scope}
\end{tikzpicture}}
  \caption{System summary graph of the DGP}
  \label{fig:dgp_summary_graph}
\end{figure}
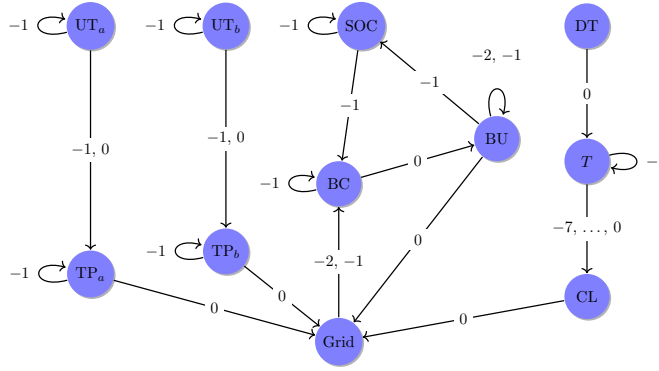
\begin{figure}
  \centering
  \adjustbox{width=\textwidth}{\input{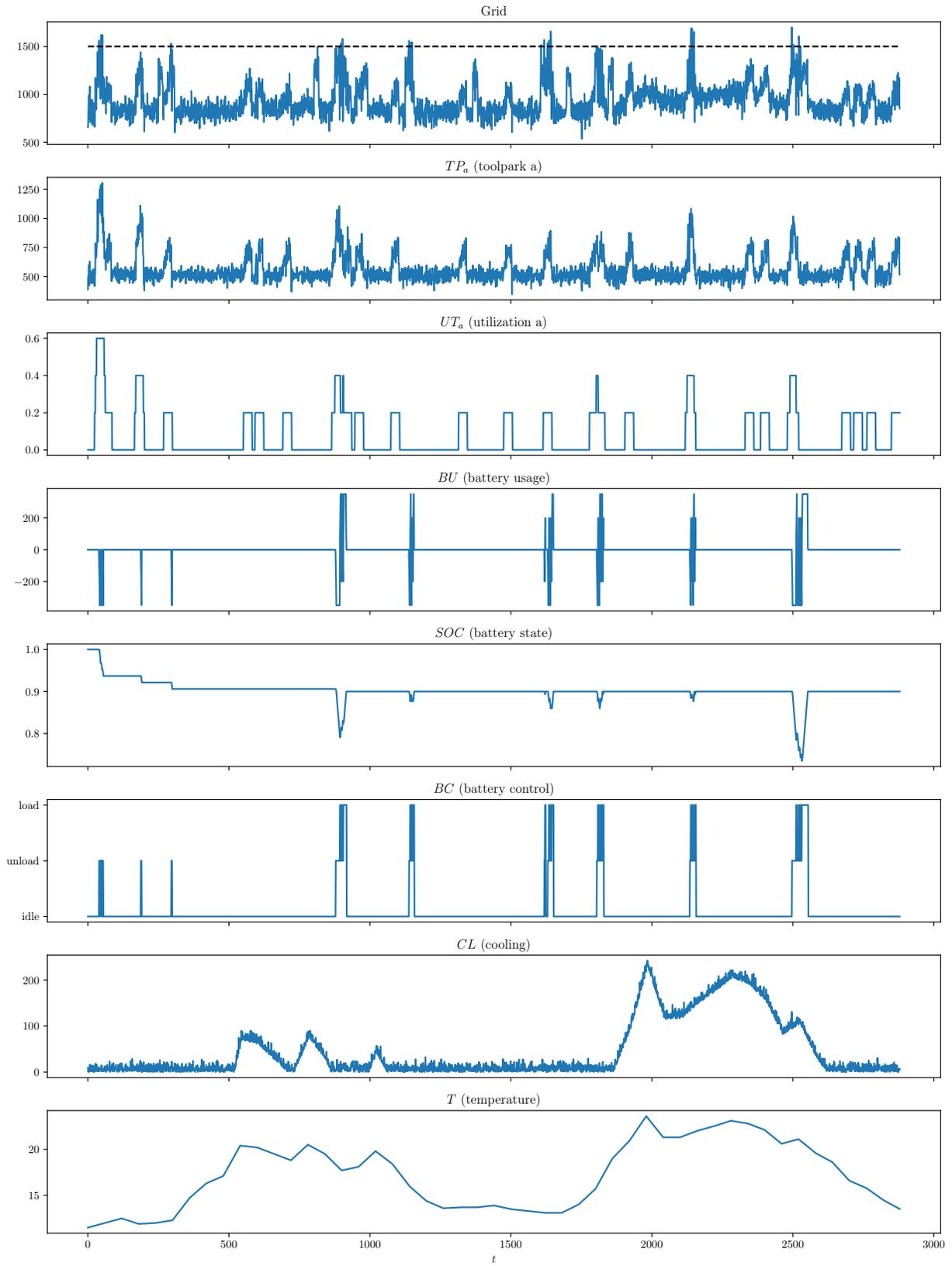}}
  \caption[Exemplary DGP output]{
    Exemplary DGP output: The dotted horizontal line in the \(\grid\) graph signifies the peak limit
  }
  \label{fig:dgp_data}
\end{figure}
Note that the used dependency graph only approximates the simulation,
since individual machine states are not included but approximated through the derived 
variables \(\util_a\) and \(\util_b\).
Moreover, the relation of temperature \(T\) and daytime \(\daytime\) is implicit since real-world data is used for these nodes.
Put differently, the simulation does not directly 
fit the functional form of Equation \ref{eq:scm_functional}
given the dependency graph in Figure \ref{fig:dgp_summary_graph}.
It rather represents a reasonable approximation,
as required in many real application settings.

\FloatBarrier
\subsection{Injections}\label{sec:injections}
\begin{figure}[t]
  \centering
  \begin{subfigure}{0.65\linewidth}
    \adjustbox{width=\textwidth}{\input{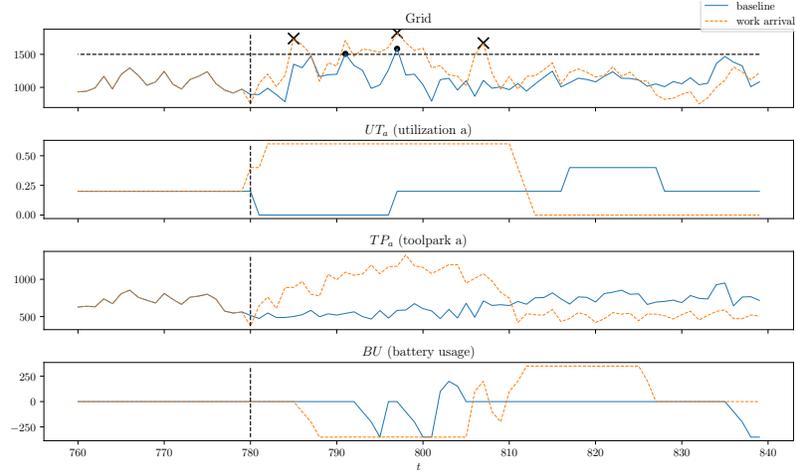}}
    \caption{work arrival}
    \label{fig:injection_work_arrival}
  \end{subfigure}%
  \\
  \begin{subfigure}{0.65\linewidth}
    \adjustbox{width=\textwidth}{\input{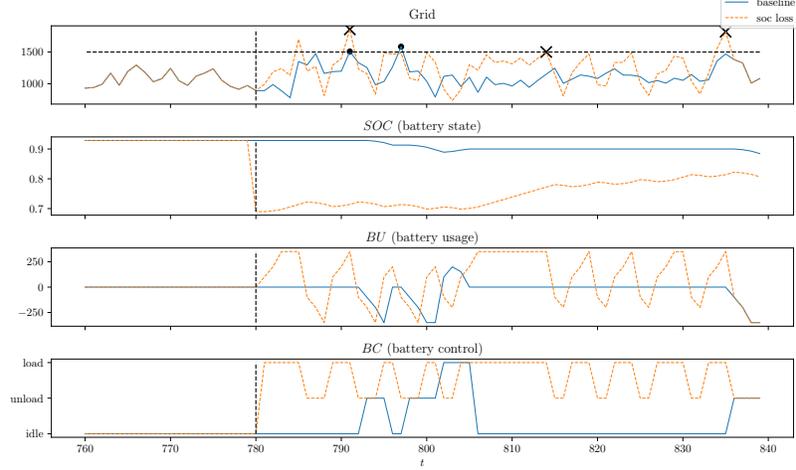}}
    \caption{soc loss}
    \label{fig:injection_soc_loss}
  \end{subfigure}\\
  \begin{subfigure}{0.65\textwidth}
    \adjustbox{width=\textwidth}{\input{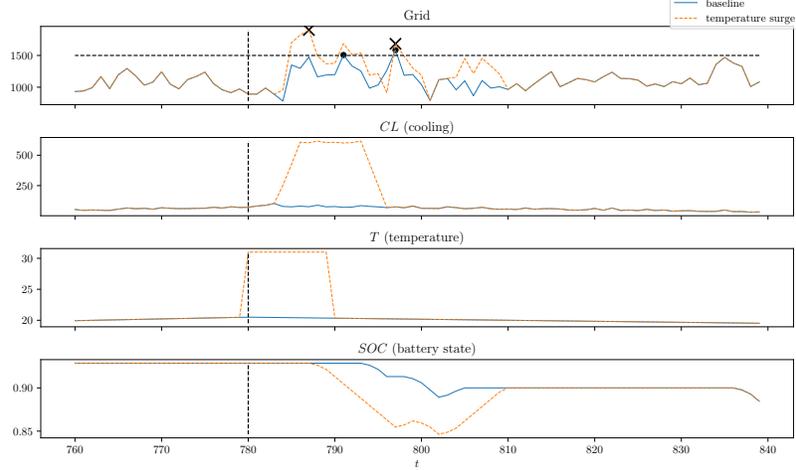}}
    \caption{temperature surge}
    \label{fig:injection_temp_surge}
  \end{subfigure}
  \caption[Exemplary system propagation with (orange) and without (blue) injection]{
    Exemplary system propagation with (orange) and without (blue) injection.
    The most relevant system nodes are shown. The injection starts at the vertical dotted line
    and the peak limit is given by the horizontal dotted line.
    Peaks caused by the injection (marked with \(X\)) are determined by comparison with the baseline.
  }
  \label{fig:injection}
\end{figure}

In this section we describe the method used to inject deviations
that lead to energy peaks with assignable root-causes.
In general two simulation runs with the same initial condition and random seeding are compared.
While the first run is not modified and represents the baseline,
the second run diverges from the first through one of the following injections:
\begin{enumerate}
  \item \textit{temperature surge}:
    The temperature is set to a high value 
    (\(T=31\degree C\))
    for an extended time period
    (\(\Delta t = 10\)), directly affecting \(T\).
    In application such a scenario may happen due to sensor failures,
    e.g. the sensors being blocked by an object, or faulty readouts.

  \item \textit{cooling surge}:
    The required cooling power is set to high value (\(\cooling = 265 kW\)) for an extended time period
    (\(\Delta t = 10\)), directly affecting \(\cooling\).
    In application this scenario may occur due to control failures.
    For example suppose that an individual cooling machine failed during startup.
    Thus, an additional machine is immediately started to compensate the required cooling power. 
    But the first machine might still draw a substantial amount of power until it is shut down.

  \item \textit{cooling scale}:
    The scaling factor used to calculate the required cooling power given the temperature
    is increased (\(75 \frac{kW}{\degree C}\)) compared to the normal setting
    (\(60 \frac{kW}{\degree C}\)) for an extended time period (\(\Delta t = 10\)),
    directly affecting the calculation of \(\cooling\).
    In an application setting the internal control target,
    that regulates the individual cooling machines might be set 
    to a high value due to sensor failures or situations of high 
    humidity might lead to a higher consumption of the cooling equipment.

  \item \textit{bat fail}:
    The battery output is set to zero (\(\batusage = 0 kW\))
    for an extended time period (\(\Delta t = 15\)),
    directly affecting \(\batusage\).
    This simulates a failure in the battery system
    in which the battery does not produce any output independently of the controller.

  \item \textit{soc loss}:
    The state of charge of the battery is immediately set to a low value (\(\batsoc = 0.69\))
    directly affecting \(\batsoc\) and causing the battery controller to load the battery
    until it is in normal \(\batsoc\) range.
    This injection has long-term effects on the system propagation, as it 
    changes the load cycles of the battery.
    Loosing a large amount of charge in a real-world application would indicate a severe failure
    of the battery system.
    We regard this as a rather uncommon but interesting case as it showcases typical
    behavior of systems with memory. 

  \item \textit{work arrival}:
    Three work items arrive at tool park \(a\) within three minutes
    affecting \(\util_a\).
    This causes several machines to be activated in a short time.
    In applications such a scenario may happen after work breaks or during a shift handover
    and may lead to peaks which would be avoidable with better synchronization.

  \item \textit{grid noise}:
    For an extended period of time (\(\Delta t = 10\)) the
    normal distributed noise variable affecting the \(\grid\) node is replaced by a
    normal distributed variable with higher standard deviation.
    The high value is given by \(\sigma = 550 kW\)
    compared to the normal setting with \(\sigma = 30 kW\).
    This behavior mimics the activation of additional consumers or producers in the plant
    that are not explicitly tracked through any of the tool parks.
    This injection is a standard non-time-dependent system alteration similar
    to the alterations used by \cite{Budhathoki22a}.
    
\end{enumerate}
In the second run the selected injection is activated at time \(t_I\)
and remains \(\Delta_I\) minutes active as specified.
In particular \(t_I\) is chosen such that the initial conditions of the simulation have faded off.
Now if a peak occurs in the second simulation run that has not been present in the baseline
simulation or is of higher magnitude than a peak in the baseline,
this peak is attributed to the node that is affected by the injection.
All other peaks are rejected.
We restrict our attention to peaks that emerge in the time interval \([t_I,\, t_I + \tau]\)
after the injection start with \(\tau > 0\).
Thus, attributable peaks may appear during an activated injection or afterward.
Figure \ref{fig:injection} visualizes the effect of the \textit{soc loss}, \textit{work arrival} and \textit{temperature surge} injections.
As it can be seen by the third injection peak in Figure \ref{fig:injection_work_arrival} the cause effect chain
spreads out through time and multiple system nodes.
The higher tool park utilization caused the battery controller to unload 
and later to load the battery again resulting in the peak.
Further, Figure \ref{fig:injection_temp_surge} showcases a time-delayed cause effect
chain where the system recovers smoothly from injection transitioning back into its baseline state after some time.
This stays in contrast to the far-reaching \textit{soc loss} injection that causes peaks with great delay as shown in Figure \ref{fig:injection_soc_loss}.

\FloatBarrier
\subsection{Benchmark}
In this section we describe our benchmark procedure.
We draw \(40\) baseline samples with different seeding from the DGP simulating 
the same time period \([t_s,\, t_I + \tau]\) starting at time \(t_s\).
For each of the injections described in Section \ref{sec:injections}
the baseline simulations are repeated using the same seeding but activating the injection at time \(t_I\).
Comparing the injected simulation runs with the corresponding baseline samples in the time interval
\([t_I,\, t_I + \tau]\) leads to 
the attributable peaks shown in Table \ref{tab:bench_peaks_overview}.
\begin{table}
  \caption[Number of attributable peaks per injection]{
    Number of attributable peaks per injection.
  }
  \centering
  \begin{tabular}{llr}
\toprule
injection & affected nodes & peaks \\
\midrule
bat fail & $BU$ (battery usage) & 10 \\
cooling scale & $CL$ (cooling) & 22 \\
cooling surge & $CL$ (cooling) & 10 \\
grid noise & Grid & 30 \\
soc loss & $SOC$ (battery state) & 96 \\
temperature surge & $T$ (temperature) & 69 \\
work arrival & $UT_a$ (utilization a) & 83 \\
\bottomrule
\end{tabular}

  \label{tab:bench_peaks_overview}
\end{table}
For some injections the number of generated peaks is lower than the number of samples.
Indeed, in some cases the battery is able to stabilize the system mitigating the injection.
The \textit{bat fail} injection disables this mitigation measure,
which in the first place does not cause any extraordinary power consumption by itself.
It rather reveals anomalies previously mitigated.
This means that in controlled, self-stabilizing systems an outlier is not likely
to be explained by a singular root-cause.
\begin{figure}
  \centering
  \adjustbox{width=\textwidth}{\input{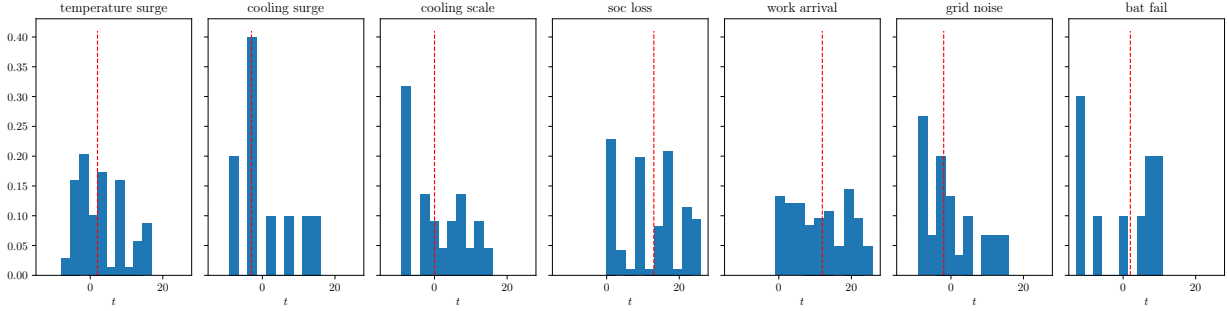}}
  \caption[Distribution of attributable peaks in time per injection]{
    Distribution of attributable peaks in time per injection:
    The time axis is relative to the injection end and
    the median time delay relative to the injection end is marked by
    the red dotted line.
  }
  \label{fig:bench_peaks_lag}
\end{figure}
Figure \ref{fig:bench_peaks_lag} visualizes the distribution of the attributable peaks in time.
The \textit{soc loss} and \textit{work arrival} peaks are more time delayed 
compared to the peaks generated through the \textit{grid noise} or \textit{cooling surge} injections.

To train the SCM we draw three samples from the DGP simulating a whole month starting from \(t_s\). 
While for each sample a different seed is used the temperature profile remains the same.
For a given maximum lag \(L\) we unfold the system summary graph shown in Figure \ref{fig:dgp_summary_graph}
as described in Section \ref{sec:methodology}.
Then causal mechanisms are assigned according to the dependencies of the unfolded graph.
The difference between the truncated and non-truncated model is that in the latter case
the dependency graph includes the dangling nodes with modified parent dependencies,
whereas the former truncates at lag \(L\) leaving the overall dependency structure intact.
At the battery control node \(\batcont\) we assign the actual control mechanisms.
This is in line with the application perspective,
where it can be assumed that the control logic is known.
This also means that \(f^{\batcont}\) is not dependent on a noise variable and thus does not get any attributions.
For the other nodes we rely on the auto assign heuristic implemented in the dowhy package \citep{Bloebaum2024,Sharma2020}.
Note that from the application perspective the system would admit adding
further physical constraints to the model.
For example, the values of \(\grid\) and its direct parents
\(\toolpark_a\), \(\toolpark_b\), \(\batusage\) and \(\cooling\) can be thought of as time synchronous
power meter measurements at their respective grid junction points.
Thus, if the measurements have the same unit, the functional model \(f^{\grid}\)
has to be linear with coefficients being equal to one.
The trained model is used to calculate attributions for each peak generated by an injection.

Further, we compare the CRCA models to a simple peak attribution heuristic (Algorithm \ref{algo:heuristic}).
The idea behind the heuristic algorithm is to explain the increase (decrease) of a node \(i\)
during the build-up of the peak through increases (decreases) of its parents in the same period
while assuming a linear structural model with a dependency tree.
The residual part of the increase is then attributed to the node \(i\) itself.
To make the different features comparable the coefficients of the linear mechanisms are used to 
reweight the attributions.
\begin{algorithm}[htb]
  \KwData{
    target limit \(y\),
    linear coefficients \(w_{i,j}\) of parents \(i\) on the children \(j\),
    data \(x\) with peak at \(x^n_t > y\),
  }
  \KwResult{attributions \(\Phi_i\) of the nodes \(i=1, \ldots, n\)}
  calculate the time lag \(s \leftarrow \min\{l \in \Nn \,|\, x^n_{t-l} > y \geq x^n_{t-l-1} \}\)
  of the first overshoot before the peak\;
  \For{all nodes \(i\)}{
    \(\delta^i \leftarrow x^i_t - x^i_{t-s}\) calculate the peak increase\;
    find the shortest path \((i,\, j_1,\, j_2,\, \ldots,\, j_k, \, n)\) from \(i\) to \(n\) in the dependency tree\;
    \(\lambda^i \leftarrow w_{i,j_1} w_{j_1,j_2} \cdots w_{j_k, n}\)\;
    reweight the increase \(\delta^i \leftarrow \delta^i \lambda^i\)\; 
  }
  \For{all nodes \(i\)}{
    determine the set of parent nodes \(\Pa_i\) of \(i\)\;
    \(\Phi_i \leftarrow \delta^i - \sum_{j \in \Pa^i} \delta^j\)\;
  }
  \caption{
    Heuristic (residual peak attributions)
  }
  \label{algo:heuristic}
\end{algorithm}
If we disregard the \(\batsoc\), \(\batcont\) and \(\daytime\) nodes as well as the 
time-lagged dependencies in the system summary graph (Figure \ref{fig:dgp_summary_graph})
the remaining dependency tree can be plausibly approximated through linear mechanisms.
As argued above the mechanism \(f^{\grid}\) is linear in its direct parents with coefficients equal to one.
The remaining coefficients are regressed from data.
This, coefficients are interpreted as average conversion factors for the increases,
e.g. in case of \(\util_a \to \toolpark_a\) the related coefficient \(w_{\util_a, \toolpark_a}\)
is understood as the average increase of cooling power per increase of the utilization.

\FloatBarrier
\subsection{Evaluation}
Denote with \(\mathcal{P}_I\) the set of peaks caused by injection \(I\)
and let \(c_I \in \{1,\, \ldots,\, n\}\) be the node affected by \(I\).
For a peak \(p \in \mathcal{P}_I\) at time \(t_p\) we
aggregate the CRCA attributions of a particular lag \(L\) model
in time to assess the feature localization capability.
Particularly, we evaluate
\[ \Phi_{p}(i) \defeq \Phi_{p}(i,\, L) \defeq \agg_{l \in \mathcal{U}_i} \phi_{t_p}(i,\, l) \]
for different lag aggregations
\(\agg \in \left\{\max, \, \sum\right\}\)
to rank the graph nodes \(i \in \{1, \ldots, n\}\).
Thereby
\(\mathcal{U}_i \defeq \{l \,|\, (i,\,l) \in \mathcal{U} \}\) 
is defined as the set of available lags for node \(i\) given the lag \(L\) model.
Then the hit rates
\[{\hit}k(I,\, L) \defeq \frac{1}{|\mathcal{P}_I|} \sum_{p \in \mathcal{P}_I} \ind[c_I \in \rank^k(p,\, L)] \]
given injection \(I\) are calculated, where
\(\rank^k(p,\, L)\) denotes the \(k\) highest ranking nodes with respect to the attributions
\(\Phi_{p}(1), \ldots, \Phi_{p}(n)\).
While the injection dependent hit rate reveals performance differences between the anomaly types,
the independent hit rate \({\hit}k(L)\) is used to equate the overall performance.
The peak heuristic attributions are evaluated accordingly without the prior aggregate over the time-lags.

To assess the time localization capability of the CRCA method we compute the predicted anomaly time
\[\hat{t}_I \defeq t_p - \argmax_{l \in \mathcal{U}_{c_I}} \phi_{t_p}(c_I,\, l)\]
as the peak time \(t_p\) minus the time lag of the targeted node \(c_I\) having the maximum attribution.
Thus, we assume that the feature localization is known.
Thereby we view the method to be successful if the predicted anomaly time
lies in the interval \([t_I,\, t_I + \Delta_I]\) during which the injection \(I\) was active.
Consequently, we use the following comparison function:
\begin{align*}
  d\left(t_I,\, \hat{t}_I\right) \defeq 
  \begin{cases}
    0, &\quad \hat{t}_I \in [t_I,\, t_I + \Delta_I]\\
    \hat{t}_I - (t_I + \Delta_I), &\quad \mbox{otherwise}
  \end{cases}
\end{align*}

%
\section{Results}\label{sec:results}
In this section we present the evaluation results, 
with the truncated model being evaluated for maximum lags \(L = 1, \ldots, 13\)
and the non-truncated for \(L = 1, \ldots, 10\).
We compare the feature localization capability of the truncated and non-truncated models.
Thereby, the peak heuristic serves as a semi-time-dependent lower bound.
By construction, it is not to be expected that the heuristic can capture root causes with long time delays
and thus should only be comparable to truncated models with low maximum lag.
Further, we evaluate the time localization performance of the CRCA models.
Lastly, the effect of the mechanism approximation in non-truncated models is 
discussed by examination of an example peak.
\FloatBarrier

\subsection{Feature Localization}
\begin{figure}[H]
  \centering
  \begin{subfigure}{1.0\linewidth}
    \adjustbox{width=\textwidth}{\input{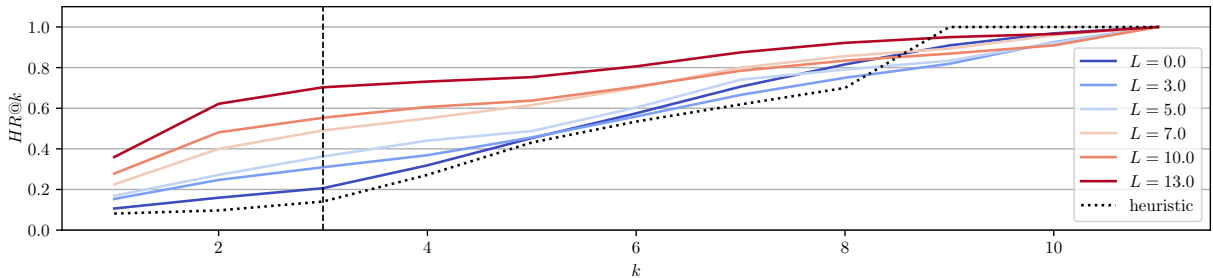}}
    \caption{Truncated models and heuristic}
    \label{fig:result:hitrate:heuristic_vs_trunc}
    \centering
  \end{subfigure}\\
  \begin{subfigure}{1.0\linewidth}
    \adjustbox{width=\textwidth}{\input{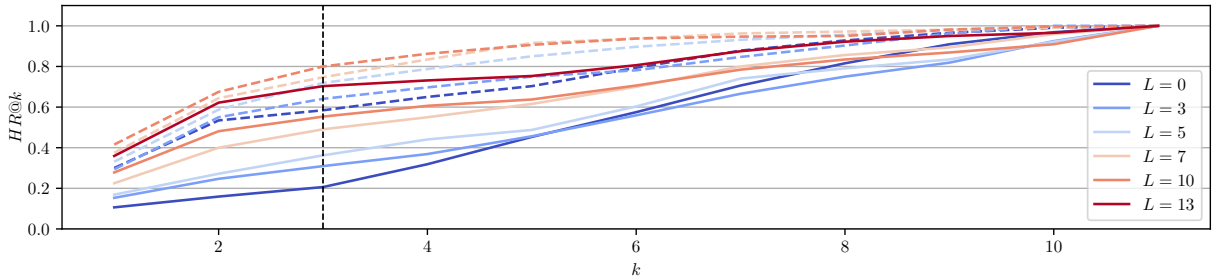}}
    \caption{Truncated (solid-lines) and non-truncated (dashed-lines) models}
    \label{fig:result:hitrate:trunc_vs_nontrunc}
    \centering
  \end{subfigure}
  \caption[\({\hit}k\) values under \(\Sigma\) aggregation]{
   \({\hit}k\) values under \(\Sigma\) aggregation.
   The vertical dashed black line represents the \({\hit}3\) values.
  }
  \label{fig:result:hitrate}
\end{figure}

As argued before, in regulated systems it is not likely that an outlier is produced by a singular cause.
Accordingly, it cannot be expected that the injected root cause is always clearly distinguished from
other potential root causes in terms of the calculated attributions.
We rather argue that the injected root cause should
at least be among the three highest ranking causes,
which makes \({\hit}3\) the main metric to compare the different variations.
Of course this depends on the number of available features.
Table \ref{tab:result:hr3_overview} gives an overview of the CRCA \({\hit}3\) results.
\begin{table}[H]
  \centering
  \caption[\({\hit}3\) overview of CRCA models]{
    \({\hit}3\) overview of CRCA models.
  }
  \begin{tabular}{lrrrr}
\toprule
 & \multicolumn{4}{r}{${\hit}3$} \\
 & \multicolumn{2}{r}{non-truncated} & \multicolumn{2}{r}{truncated} \\
$\agg$ & $\max$ & $\sum$ & $\max$ & $\sum$ \\
$L$ &  &  &  &  \\
\midrule
0 & 0.6062 & 0.5844 & 0.2031 & 0.2062 \\
3 & 0.6562 & 0.6406 & 0.3094 & 0.3094 \\
5 & 0.7250 & 0.7188 & 0.3312 & 0.3625 \\
7 & 0.8094 & 0.7469 & 0.4688 & 0.4906 \\
10 & 0.8187 & 0.8000 & 0.5437 & 0.5531 \\
11 & - & - & 0.6687 & 0.6625 \\
13 & - & - & 0.7031 & 0.7031 \\
\bottomrule
\end{tabular}

  \label{tab:result:hr3_overview}
\end{table}
The peak heuristic with a \({\hit}3\) value of \(0.1406\) is outperformed by all other variations.
This was to be expected for truncated and non-truncated models with maximum lag \(L > 0\).
But even the truncated \(L=0\) model has a slight advantage over the heuristic,
as the latter is not able to attribute a value to \(\batsoc\) and thus to
detect the \textit{soc loss} injection.
In general the hit rate of the CRCA models is increasing with increasing \(L\),
indicating that delayed anomalies might get successively captured by models with higher maximum lag.
Under the truncated models the summation aggregation leads to slightly better values
compared to the maximum aggregation.
The reverse is true for the non-truncated models.
Surprisingly the non-truncated models outperform the truncated models with respect to \({\hit}3\).
One reason is that the effective maximum lag is greater than \(L\) for non-truncated models
as they can evaluate the dangling nodes that are otherwise truncated.
The problem of the non-truncated structural model is the mechanism approximation at the dangling nodes.
It seems that in our particular case the dropping of parent dependencies at these nodes
does still lead to good enough estimates, such that the benefit of additional
lags outweighs the approximate nature of non-truncated models.
Figure \ref{fig:result:hitrate} visualizes the overall hit rate \({\hit}k\) comparing
the truncated models with the peak heuristic (Figure \ref{fig:result:hitrate:heuristic_vs_trunc})
and the truncated with the non-truncated models (Figure \ref{fig:result:hitrate:trunc_vs_nontrunc}).
Figure \ref{fig:result:hitrate:trunc_vs_nontrunc} shows that
only above maximum lag \(L=10\) the truncated models are in the range of the non-truncated.

\begin{figure}
  \centering
  \adjustbox{width=\textwidth}{\input{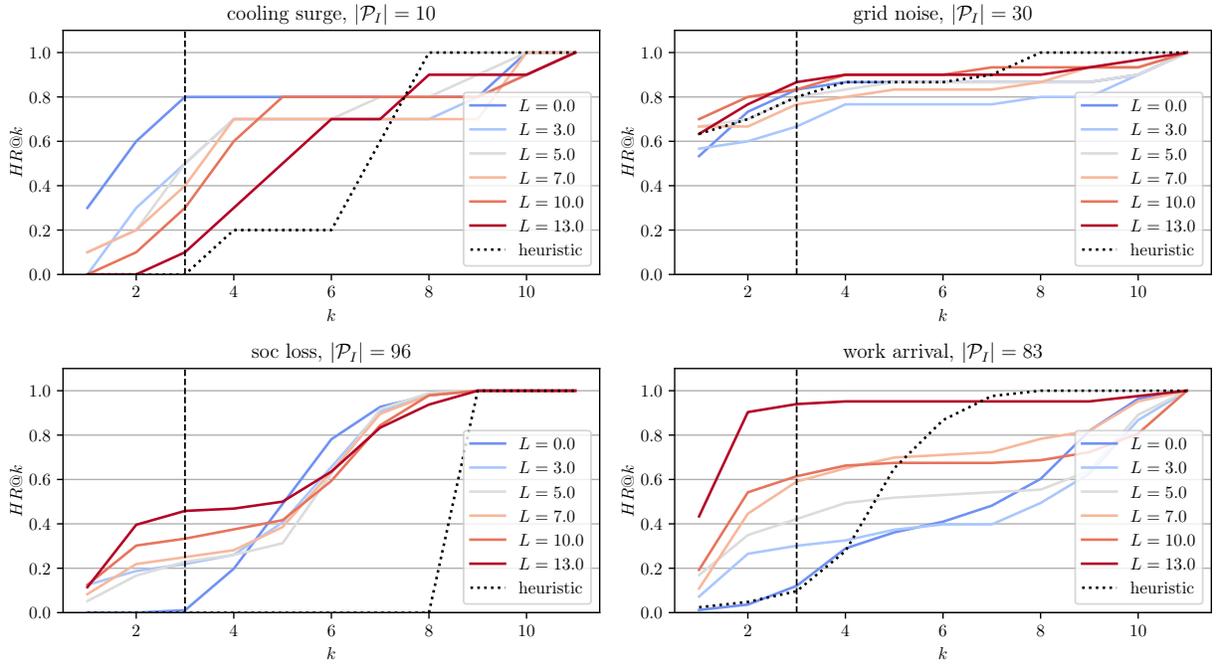}}
  \caption[Hit rate \({\hit}k\) per injection under \(\Sigma\) aggregation of heuristic and truncated models]{
    Hit rate \({\hit}k\) per injection under \(\Sigma\) aggregation of heuristic and truncated models:
    The hit rate of \textit{soc loss} and \textit{work arrival} injections increase with \(L\)
    (Category \ref{cat:increase})
    while for \textit{grid noise} there is little benefit in additional lags
    (Category \ref{cat:lag_independent}).
    For \textit{cooling surge} increasing lags seem non-beneficial (Category \ref{cat:no_pattern}).
  }
  \label{fig:result:hitrate:heuristic_vs_trunc_inj_exemplary}
\end{figure}

Before investigating this effect in more detail we are going to evaluate
the hit rate dependence on the maximum lag per injection.
We can observe three basic patterns depending on the complexity of the injection:
\begin{enumerate}
  \item\label{cat:increase} Increasing with \(L\)

    The hit rate increases substantially with an increase of the maximum lag \(L\).
    These injection types are better captured by models with high maximum lag.

  \item\label{cat:lag_independent} Little benefit of higher maximum lags \(L\)

    The hit rate is on a high level independently of the maximum lag \(L\).
    That is, there is no significant gain or loss switching from a model with 
    low to a model with high maximum lag.

  \item\label{cat:no_pattern} Non-beneficial / No pattern in \(L\)

    While there are significant differences between the models with different
    \(L\) we can not observe a substantial increase of the hit rate with an increase of \(L\).
\end{enumerate}
An overview of the different patterns that emerge using the truncated model 
is provided by Figure \ref{fig:result:hitrate:heuristic_vs_trunc_inj_exemplary}
and the full comparison of all injections can be found in Appendix \ref{app:results}
in Figure \ref{fig:result:hitrate:heuristic_vs_trunc_inj_full}.
For the truncated model we observe that
the \textit{temperature surge}, \textit{work arrival},
\textit{soc loss} and \textit{bat fail} injections
fall into the increasing category (Category \ref{cat:increase}).
These injections target parts of the system that have a time delayed effect on the target node,
which explains the observation.
While the hit rate under the \textit{soc loss} peaks increases too with \(L\),
the detection rate remains low.
This not only lies in the long ranging effect of the injection, 
which is comparable to \textit{work arrival},
but also in its indirect nature that prevents compensation of peaks.
Also, the \textit{bat fail} injection is increasing with \(L\), 
except for individual lags (e.g. \(L=13\)).
Since its effect is rather immediate the differences between the 
different maximum lags are not as prominent.
The low number of samples might also not allow the observation of a clear trend.
Note that a disabled battery does not cause peaks by itself.
It rather prevents peaks that would have been caused by other consumption nodes.
The detection of \textit{grid noise} and \textit{cooling scale} peaks
do not benefit from an increase in \(L\) (Category \ref{cat:lag_independent}).
The \textit{grid noise} injection has an immediate effect on the target node
and its structure is well aligned with the model structure.
As we argued before this type of noise modification is an obvious change
to be made in causal structural models.
The injection produces immediate distinct peaks such that
even the heuristic is able to outperform the truncated \(L=0\) model.
In comparison, the \textit{cooling scale} injection is more complex
having a greater influence on the battery system
potentially causing \(\grid\) peaks through delayed load cycles.
The low performance of the peak heuristic can be explained 
through the constant factor affecting the \(\cooling\) increase during the injection time.
Thus, there might hardly be any change in \(\cooling\) during the peak build up,
leading the peak heuristic to assign a low attribution.
The \textit{cooling surge} injection falls into the remaining category (Category \ref{cat:no_pattern}),
as we observe a decreasing hit rate with increasing \(L\).
This observation might be a result of the low number of sample peaks.
The injection design could provide another explanation.
In contrast to \textit{cooling scale} a fixed high consumption value is injected in \(\cooling\).
This is supposed to result in a difference compared to the baseline sample.
However, in the observed time period this difference is rather subtle
compared to the difference caused by the scale-up of \textit{cooling scale}.
This leads to peaks in which the \(\cooling\) node behaves less distinctive
and thus to the difficulty to recover \(\cooling\) as the root cause.

We now investigate how the categorization changes in the non-truncated case.
An overview is provided with Figure \ref{fig:result:hitrate:nontrunc_vs_trunc_inj_exemplary}
and a full comparison can be found in Appendix \ref{app:results} with
Figure \ref{fig:result:hitrate:nontrunc_vs_trunc_inj_full}.
\begin{figure}
  \centering
  \adjustbox{width=\textwidth}{\input{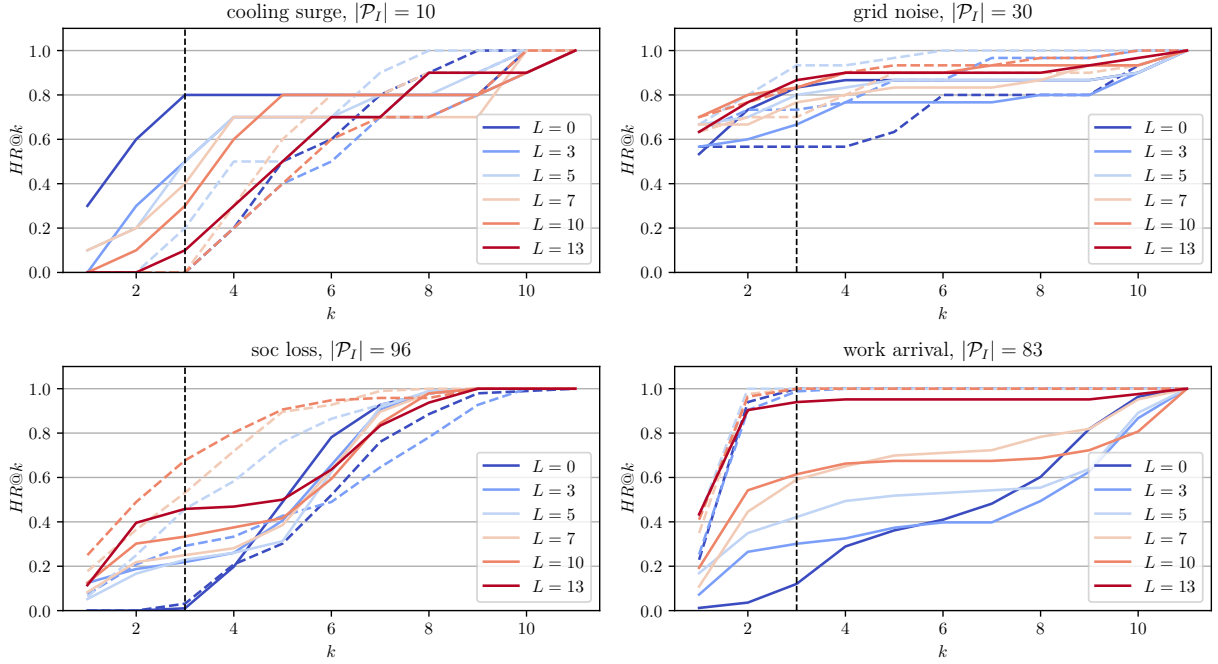}}
  \caption[Hit rate \({\hit}k\) per injection under \(\Sigma\) aggregation of truncated and non-truncated models]{
    Hit rate \({\hit}k\) per injection under \(\Sigma\) aggregation of
    truncated (solid lines) and non-truncated models (dashed lines):
    The hit rate of \textit{soc loss} remains increasing with \(L\)
    in the non-truncated case (Category \ref{cat:increase}) outperforming the truncated models
    given high enough \(L\).
    As in the truncated case variations of \(L\) 
    have only little effect on the already high hit rate 
    of \textit{grid noise} injections (Category \ref{cat:lag_independent}).
    Both methods are on par in this case. 
    While non-truncated models perform worse for \textit{cooling surge} peaks
    (Category \ref{cat:no_pattern}),
    they surpass the truncated models for \textit{work arrival} injections
    maintaining a high hit rate independent of \(L\)
    (Category \ref{cat:lag_independent}).
  }
  \label{fig:result:hitrate:nontrunc_vs_trunc_inj_exemplary}
\end{figure}
For injections with time-delayed effects we see the most improvement compared to the truncated models.
In particular for \textit{soc loss}, \textit{temp surge} and \textit{work arrival}
peaks the non-truncated models outperform the truncated.
For the first two we observe an increase of the hit rate with increasing \(L\) as before
(Category \ref{cat:increase}). 
\textit{Work arrival} peaks are even independently of \(L\) detected with high hit rate
(Category \ref{cat:lag_independent}).
Only the \(L=13\) truncated model comes close to hit rates of the truncated models 
in case of \textit{temperature surge} and \textit{work arrival} injections.
This observation is still present but not as distinct for \textit{soc loss} peaks.
For \textit{grid noise} peaks we can see the same lag independent pattern as before
(Category \ref{cat:lag_independent}). 
With exception to \(L=0\) both methods achieve comparable results for this injection.
In case of the \textit{bat fail}, \textit{cooling surge} and \textit{cooling scale}
injections we do not see distinct increases with \(L\) or a lag independent high hit rate
(Category \ref{cat:no_pattern}).
While the non-truncated models performs worse for the cooling injections
they yield a comparable hit rate performance for the \textit{bat fail} injection.
Again a higher number of peak samples may allow drawing different conclusions for \textit{bat fail}
and \textit{cooling surge}.
In general these observations support the hypothesis that
the advantage of the non-truncated model is due to its higher effective lag.

\begin{figure}
  \centering
  \adjustbox{width=\textwidth}{\input{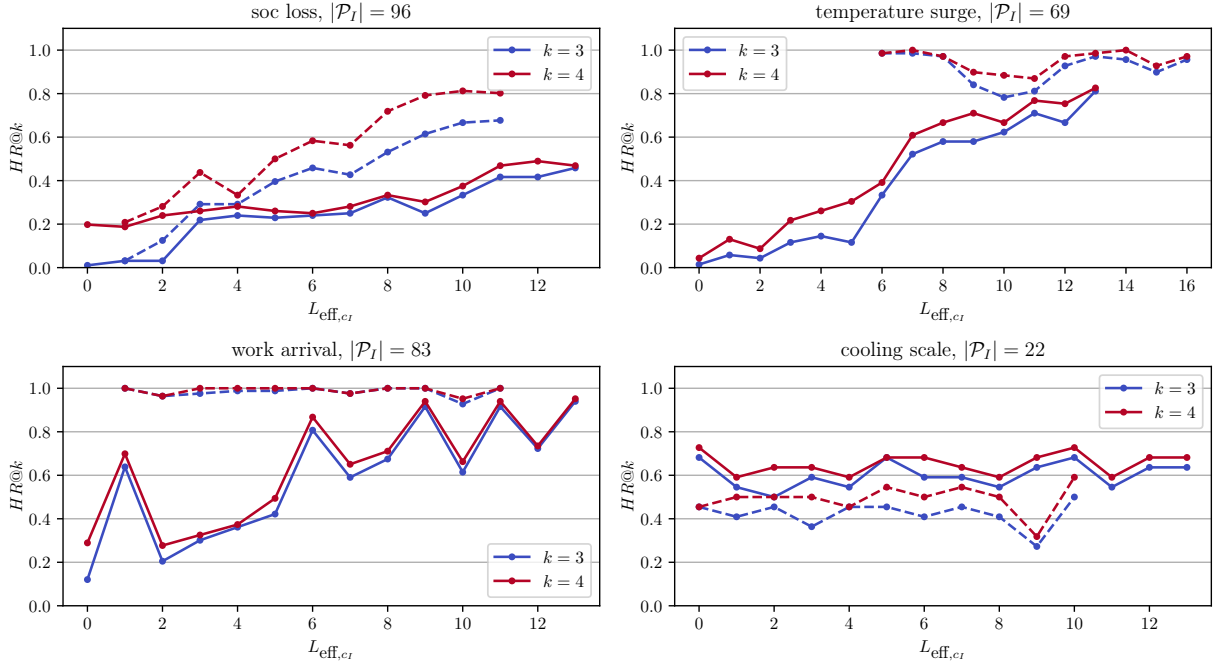}}
  \caption[\({\hit}k\) under the \(\Sigma\) aggregation in dependence of the effective maximum lag]{
    \({\hit}k\) under the \(\Sigma\)
    aggregation in dependence of the effective maximum lag \(L_{\eff, c_I}\)
    of the truncated (solid line) and non-truncated (dashed line) models.
  }
  \label{fig:result:hitrate:effectiv_lag_inj_exemplary}
\end{figure}

As the hit rate comparison of truncated and non-truncated models with same maximum lag \(L\)
is biased towards the non-truncated models in terms of attributable nodes
we now compare both using their effective maximum lag.
Given a node \(i\) we define the effective maximum lag
\(L_{\eff, i} \defeq \max_{i} \mathcal{U}_{i}\)
to be the maximum attributable lag of the node \(i\).
We then use \(L_{\eff, c_I}\) to compare the two approaches on an injection
basis with \(c_I\) being the node affected by the injection \(I\).
As can be seen in Figure \ref{fig:result:hitrate:effectiv_lag_inj_exemplary}
this partially explains the previous observed gap between both approaches.
The closing of the gap between both approaches is most visual
for the \textit{temperature surge} injection, 
which is not surprising as the \(\cooling\) node has the highest number of time-lagged dependencies.
Also for \textit{work arrival} the truncated approach catches up more closely with increasing effective lag.
One can also see that there remains a rather high gap in the \textit{soc loss} case.
We interpret the remaining gaps as follows:
A better performance in one node also stabilizes the localization performance of the remaining
features as the attributions sum to the IT-Score according to Equation \ref{eq:decomposition}.
That is even if we compare the truncated and non-truncate approaches regarding
\textit{work arrival} peaks on basis of \(L_{\eff, \toolpark_a}\) 
the non-truncated model still might benefit from
\[L_{\eff, T} > L_{\eff, \toolpark_a}\]
for the detection of anomalies in \(\toolpark_a\).
For a comparison of all injections we refer to Figure \ref{fig:result:hitrate:effectiv_lag_inj_full}
in Appendix \ref{app:results}.

\FloatBarrier
\subsection{Time Localization}
\begin{figure}[H]
  \centering
  \adjustbox{width=\textwidth}{\input{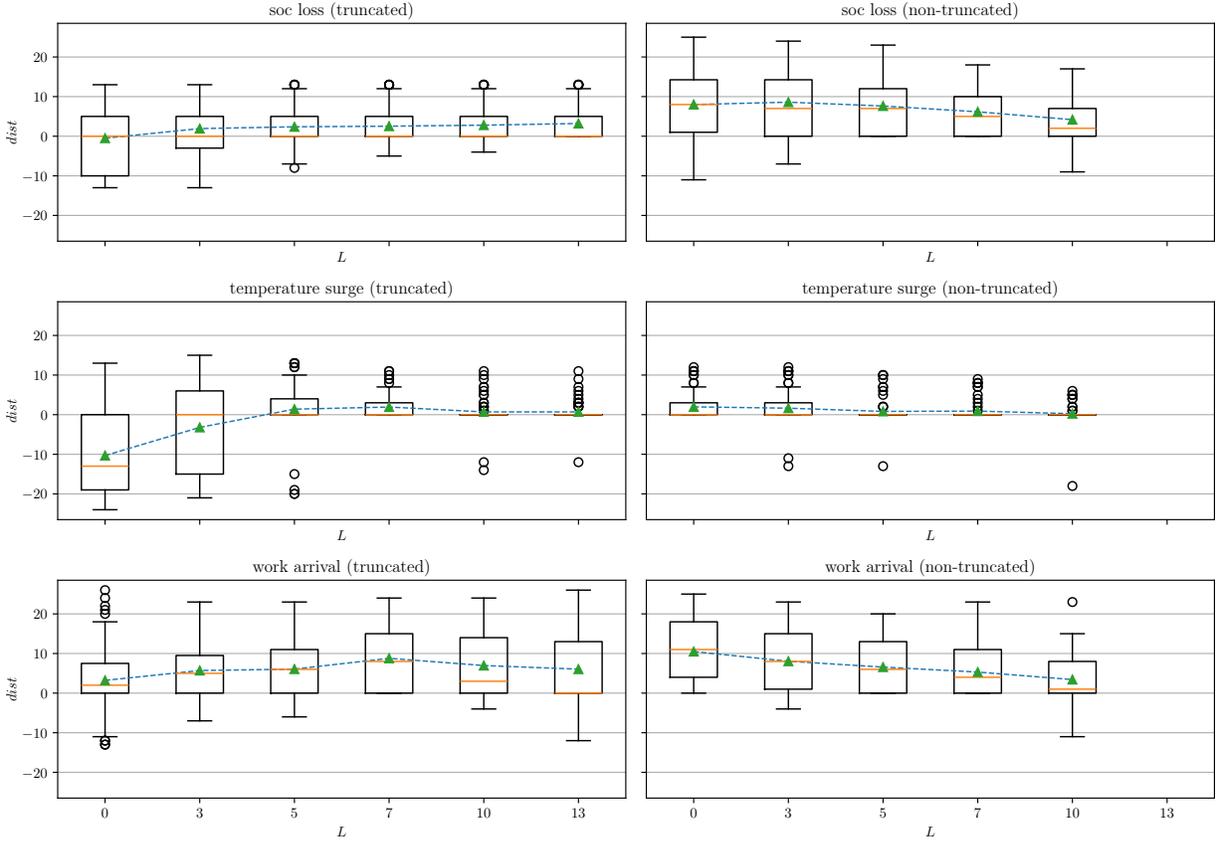}}
  \caption[Box plots of differences \(d(t_i,\, \hat{t}_i)\) among all peaks caused by delayed injections]{
    Box plots of differences \(d(t_i,\, \hat{t}_i)\) among all peaks 
    caused by delayed injections for truncated and non-truncated models.
    The mean difference is given by the dashed blue line.
  }
  \label{fig:result:time_dist:time_delayed}
\end{figure}

In this section we evaluate the time localization capability in dependence of \(L\)
for the truncated and non-truncated approaches.
Table \ref{tab:result:time_dist_overview} provides an overview
of the mean difference \(d(t_I,\, \hat{t}_I)\) between actual and predicted injection time.
Averages over all peaks are put in relation
to averages over peaks within the range of the \(L=10\) models.
The latter means that only peaks with a delay
\(\Delta_p \defeq t_p - (t_I + \Delta_I)\) of less than \(10\) time steps are considered.
Over all peaks and injections the mean difference decreases with \(L\) for the non-truncated
models. Thereby the injection time \(\hat{t}_I\) is on average estimated too high,
indicating that models with higher \(L\) might further improve the estimates.
In the truncated case models with \(L \leq 5\) seem to underestimate \(t_I\)
(real injection time is after the predicted), 
resulting the lower average compared to the non-truncated.
The truncated models start to improve with the inclusion of lags greater than six time steps.
This hypothesis is backed by the corresponding averages for peaks with delay \(\Delta_p \leq 10\).
Under this restriction the averages for both, the truncated and the non-truncated models,
are getting closer to a zero mean with higher \(L\).
Thereby truncated model exhibits a better average time localization behavior for \(L \geq 7\).
Note that after this threshold the feature location capability of the truncated models are also improving
as they start to capture injections with time delay
(see Figure \ref{fig:result:hitrate:heuristic_vs_trunc} and \ref{fig:result:hitrate:effectiv_lag_inj_full}).
In particular only for \(L>6\) the truncated model can attribute to all the lagged temperature dependencies
(see DGP in Figure \ref{fig:dgp_summary_graph}).
Further, the advantage of the non-truncated model for unrestricted \(\Delta_p\)
is not surprising given its additional attributable nodes.
\begin{table}
  \centering
  \caption[Average difference \(d(t_p,\, \hat{t}_p)\) of predicted and actual injection time per maximum lag \(L\)]{
    Average difference \(d(t_p,\, \hat{t}_p)\) of predicted and actual injection time
    per maximum lag \(L\).
  }
  \begin{tabular}{lrrrr}
\toprule
 & \multicolumn{4}{r}{$d(t_p,\, \hat{t}_p)$} \\
 & non-truncated & truncated & non-truncated & truncated \\
$\Delta_p$ & $\leq 10$ & $\leq 10$ & $< \infty$ & $< \infty$ \\
$L$ &  &  &  &  \\
\midrule
0 & 0.4779 & -7.2059 & 5.9781 & -1.1438 \\
3 & 0.1103 & -2.7647 & 5.5187 & 1.9750 \\
5 & 0.1029 & -0.6544 & 4.6688 & 3.1562 \\
7 & 0.0441 & 0.0294 & 3.9531 & 3.9875 \\
10 & -0.3897 & -0.2500 & 2.6344 & 3.2344 \\
13 & - & -0.1765 & - & 3.0125 \\
\bottomrule
\end{tabular}

  \label{tab:result:time_dist_overview}
\end{table}
Figure \ref{fig:result:time_dist:time_delayed} visualizes the difference distributions
for the time delayed injections \textit{soc loss}, \textit{temperature surge} and \textit{work arrival}.
In general the time localization improves with higher maximum lag \(L\) decreasing the standard deviation.
For \textit{temperature surge} under the truncated model this effect is the most striking
and can be explained like before through the dependency structure of the DGP for \(L \leq 7\). 
Note that the decrease also continues for \(L > 7\) as more injections fall into the model range.
For the \textit{work arrival} peaks we argue that the actual decrease 
only starts after the model is able to fully attribute to the temperature lags of the DGP.
That is, for \(L < 7\) we suspect that the model is more sensitive to work arrivals
as it divides the IT-Score \(S(x^n_t)\) among fewer nodes.
These nodes do not capture the whole effect temperature has on cooling,
while the contrary is true for the pronounced effect a high utilization has on the tool park consumption.
For the more immediate injections the decrease in \(L\) is still present but not as pronounced
(see Figure \ref{fig:result:time_dist:immediate} in Appendix \ref{app:results}).
An observation that changes when restricting to peaks within reach of the \(L=10\) models 
as can bee seen in Figure \ref{fig:result:time_dist:restricted} in Appendix \ref{app:results}.
Note that the results of the \textit{bat fail} and \textit{cooling surge} injection
in this regard are subject to a low sample size.
By looking at the \textit{soc loss} injection in \ref{fig:result:time_dist:restricted}
the difference between the truncated and non-truncated approach becomes clearly visible:
the former lacks attributable lagged nodes which results in a
systematic under estimation while
the latter is prone to the mechanism miss specification at dangling nodes.
Due to this the dangling nodes - which by definition have the highest lags in the model -
may receive higher attributions resulting in the over estimation.
In Section \ref{sec:example_peak} we analyze this effect in greater detail.

We conclude that the truncated model requires \(L\) to be high enough
to include all time lagged dependencies of the mechanisms
to correctly localize immediate anomalies - in our case \(L=7\).
Adding further lags then pays in on to the localization of delayed anomalies. 

\FloatBarrier
\subsection{Peak example}\label{sec:example_peak}
\begin{figure}
  \centering
  \adjustbox{width=\textwidth}{\input{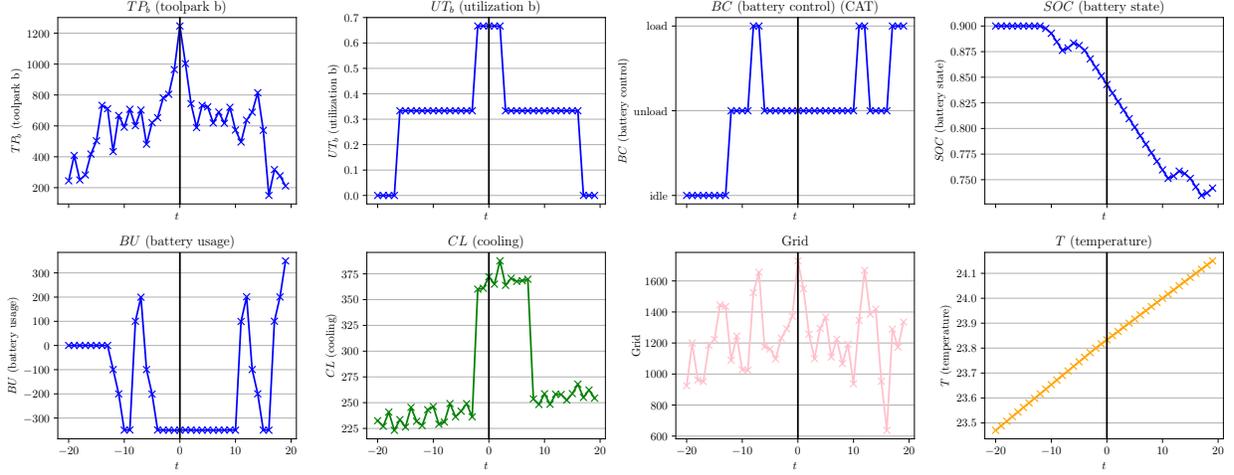}}
  \caption[Example peak caused by the \textit{cooling surge} injection]{
    Example peak caused by the \textit{cooling surge} injection: 
    In addition to the injected cooling anomaly, the tool park \(\toolpark_b\)
    has a consumption contributes to the peak in the \(\grid\) node. 
    Node regarding the other tool park are omitted as they behave normal.
  }
  \label{fig:result:cooling_surge_example:peak_detail}
\end{figure}

Lastly we investigate a specific peak caused by the \textit{cooling surge} injection in greater detail.
To inspect how the model approximation at the dangling nodes affects the prediction,
we deliberately select a peak for which the non-truncated approach has difficulties.
As shown in Figure \ref{fig:result:cooling_surge_example:peak_detail} 
the nodes \(\toolpark_b\) and \(\cooling\) are having a high power consumption
contributing to the peak observed in the \(\grid\) node.
The high tool park consumption is in line with the utilization \(\util_b\),
while the \(\cooling\) anomaly is due to the injection.
The other nodes do not exhibit exceptional behavior.
Since the peak occurs during the injection (in particular \(t_I+2 = t_p\))
models with a low maximum lag should be able to detect the anomaly.
When comparing the non-truncated \(L=1\) with the truncated \(L=2\) and \(L=7\) model
we observe that the latter do rank cooling \(\cooling\) among the three highest
root causes whereas the former does not
(see Table \ref{tab:result:cooling_surge_example:rank}).
All models correctly attribute a high value to the tool park related nodes,
but the non-truncated model wrongly attributes a high value to the temperature \(T\).
\begin{table}[H]
  \centering
  \caption[Highest ranking root causes for the example peak in Figure \ref{fig:result:cooling_surge_example:peak_detail})]{
    Highest ranking root causes of truncated and non-truncated model under \(\Sigma\)
    lag aggregation for the example peak 
    in Figure \ref{fig:result:cooling_surge_example:peak_detail})
  }
  \begin{tabular}{lllll}
\toprule
 & \multicolumn{2}{r}{non-truncated} & \multicolumn{2}{r}{truncated} \\
$L$ & 1 & 2 & 2 & 7 \\
rank &  &  &  &  \\
\midrule
1 & $\toolpark_b$ & $\util_b$ & $\util_b$ & $\util_b$ \\
2 & $\util_b$ & $\cooling$ & $\toolpark_b$ & $\toolpark_b$ \\
3 & $T$ & $\toolpark_b$ & $\cooling$ & $\cooling$ \\
\bottomrule
\end{tabular}

  \label{tab:result:cooling_surge_example:rank}
\end{table}

\begin{figure}[ht]
  \centering
  \begin{subfigure}{1.0\linewidth}
    \adjustbox{width=\textwidth}{\input{figures/result_cs_trunc_attrs.pgf}}
    \caption{Attributions \(\phi_p(i, l)\)}
    \label{fig:result:cooling_sruge_example:attrs}
    \centering
  \end{subfigure}\\
  \begin{subfigure}{0.5\linewidth}
    \adjustbox{width=\textwidth}{\input{figures/result_cs_trunc_dist_temp.pgf}}
    \caption{
      Noise distributions of \(T\) at lag \(L=5\)
    }
    \label{fig:result:cooling_surge_example:temp_noise}
    \centering
  \end{subfigure}~%
  \begin{subfigure}{0.50\linewidth}
    \includegraphics[width=\textwidth]{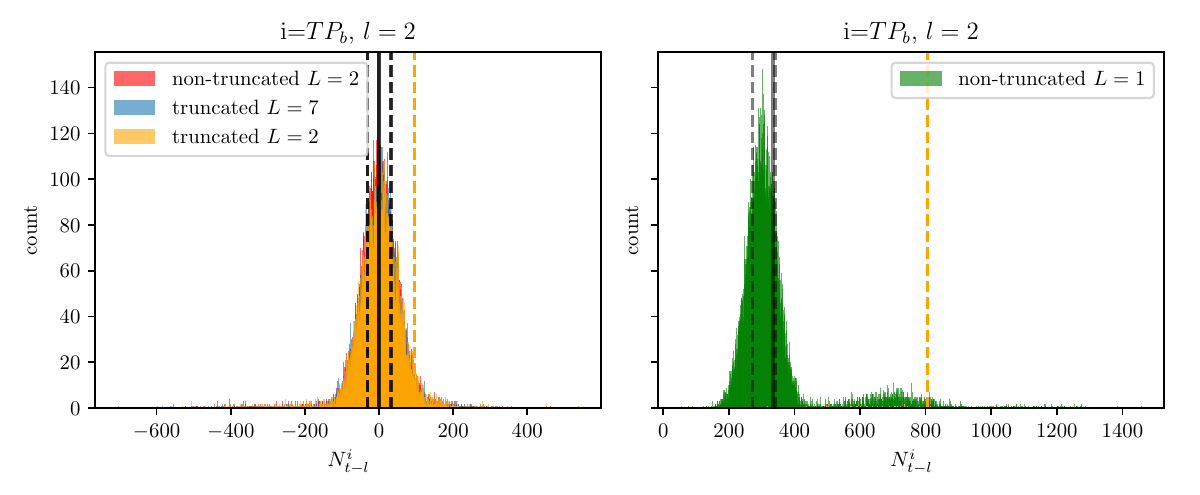}
    \caption{Noise distributions of \(\toolpark_b\) at lag \(L=2\)}
    \label{fig:result:cooling_surge_example:util_noise}
    \centering
  \end{subfigure}
  \caption[Time dependent attributions and noise distributions for the example peak from Figure \ref{fig:result:cooling_surge_example:peak_detail}]{
    Time dependent attributions and noise distributions for the example peak from Figure
    \ref{fig:result:cooling_surge_example:peak_detail}:
    The divergent attributions of the truncated and non-truncated models
    at \(l=2\) for \(\toolpark_b\) and at \(l=5\) for \(T\)
    (Figure \ref{fig:result:cooling_sruge_example:attrs})
    are related to the noise distributions of the models
    (Figures \ref{fig:result:cooling_surge_example:temp_noise} and 
    \ref{fig:result:cooling_surge_example:util_noise}):
    The orange vertical dashed lines in the distribution 
    figures signify the observed noise values of the peak at the corresponding time-lagged node.
    The \(q_{0.25}\) and \(q_{0.75}\) quantiles are given by black vertical dashed lines.
    In the long-tailed distributions of the non-truncated \(L=1\) model
    the observed noises are perceived as exceptional.
    This stays in contrast to distributions of the same nodes for the other models
    and explains the high attributions of the non-truncated \(L=1\) model.
  }
  \label{fig:result:cooling_sruge_example:attr_detail}
\end{figure}

By inspecting the time-lagged attributions we see high attribution values
for \(\phi_p(\util_b,\, 2)\), \(\phi_p(\toolpark_b,\, 2)\) and \(\phi_p(T,\, 5)\)
under the non-truncated \(L=1\) model
(Figure \ref{fig:result:cooling_sruge_example:attrs}).
These are compensated through lower attributions values at
the \(\grid\) node and for \(\phi_p(\cooling,\, 0)\)
resulting in the detection failure.
Comparing the non-truncated \(L=1\) model with the non-truncated \(L=2\) model
we see that the attributions are shifted back assigning higher values at \(\grid\)
and for \(\phi_p(\cooling,\,0)\), and lower values for 
\(\phi_p(\util_b,\, 2)\), \(\phi_p(\toolpark_b,\, 2)\) and \(\phi_p(T,\, 5)\).
This causes the non-truncated model \(L=2\) model to identify the injected anomaly
(see Table \ref{tab:result:cooling_surge_example:rank}).
However, the comparable truncated models 
attribute more steady and lower absolute values to the \(\grid\),
the \(T\) and \(\toolpark_b\) nodes at \(l \in \{2,\, 5\}\).
This leads to more pronounced attributions for \(\util_b\) and \(\cooling\).
Investigating the noise distributions of the temperature
\(N^T_{t-5}\) and tool park consumption \(N^{\toolpark_b}_{t-2}\) we see the reason for this:
the distributions of the non-truncated \(L=2\) model have a longer tail
as the corresponding functional model does not include all dependencies.
This is shown in the Figures
\ref{fig:result:cooling_surge_example:temp_noise}
and \ref{fig:result:cooling_surge_example:util_noise}.
Due to this mechanism approximation the observed noise terms,
depicted by the dashed orange line, are perceived as more exceptional
and leading to higher attributions in contrast to models with intact mechanisms.
Indeed, the tool park consumption distribution \(N^{\toolpark_b}_{t-2}\)
of the non-truncated \(L=2\) and the truncated models does not differ
as the mechanism of the former is not misspecified.
As this approximation effect can not be isolated,
we conclude that it is important to keep all attributable mechanisms intact.

\section{Conclusion}
In this paper we adapted the CRCA method of \cite{Budhathoki22a} to time dependent systems.
Our method can be interpreted as an extension of the causal formalization
of the term ``root-cause'' given by \cite{Budhathoki22a} to time-lagged systems.

We studied two methods that transform the infinite dependency graph
of time-dependent systems into a finite one, a property required to apply CRCA.
The truncation approach ensures that the causal mechanisms of all
nodes up to a maximum time-lag are left intact
by conditioning on direct parent dependencies that exceed this maximum.
While this is the obvious thing to do from a causal perspective,
it limits the model range to exactly this maximum lag.
The non-truncation approach approximates the mechanism
of the nodes conditioned by removing their dependencies.
This results in a higher number of attributable nodes and 
thus potentially to a higher time range at the sacrifice
of a causally correct model specification.
We benchmarked both approaches using a DGP
related to energy management in manufacturing systems.
The DGP is memory afflicted, has a control mechanism that affects the system target,
and exhibits instantaneous and time-delayed effects. 
All these properties make it challenging to recover root-causes.
Moreover, we argue that our work provides a blueprint for 
an application of the CRCA method in energy systems 
in order to explain peaks in the power consumption peaks.

Our results show that the CRCA adaption is able to recover
injected root-causes with different delays broadening the scope of CRCA.
Surprisingly, the non-truncation outperforms the truncation approach in the feature localization,
whereas for time-localization the latter performs more stable 
for causes in the range of the model.
Our analysis indicates that truncation prevents an over attribution at the start nodes,
at the same time limiting the model range.
That is, it prevents that wrong nodes get a high attribution
at the expense of the actual root-cause being out of the model range.
In our particular case this effect seems to payoff in favor of the non-truncated model.
Nevertheless, this might not be a general tendency as the dependency structure
of the investigated system might be more prone to confounding,
and thus invalidating the approximations made in the non-truncation case even more.
In case of the truncation approach we argue that at least all time lagged dependencies
of the mechanisms should be attributable to consistently recover instantaneous root-causes.
Every further lag then expands the actual model range.

Like for the time independent CRCA method the downside of the methodology 
is the required computational effort to evaluate the Shapley values.
Under the given independence assumptions an efficient implementation
can limit the memory complexity to the non-time dependent case and
at least reduce the computational complexity for the noise sampling and model estimation
but not for the Shapley symmetrization.
The necessary computational improvement of the method is still subject to further research.
Some recent approaches try to improve the method in case of distinct singular
root-causes \cite{Okati2024}, a scenario that we deem unrealistic for controlled systems.
In the time-dependent domain recurrent approaches,
e.g. a LSTM like causal automata that scans the time series and produces consequent attributions
based on its own memory could be a promising future research direction in this regard.
Moreover, in our specific application scenario we asked of
which causes had lead to the emergence of a peak in the energy consumption.
To be precise, we only gave answer to half of it. 
Namely, which factors had contributed to a high consumption.
But the nature of the peak is one that has a symmetry requiring to also ask:
what were the factors that had led to a decline after the peaks maximum?
As this ``going back to normal'' can also be regarded as an outlier event
the same methodology might be used to give an answer to this.
While we focused to benchmark a single method on a challenging DGP,
further research should also benchmark the pure causal CRCA approach
against the time-dependent granger causality approaches.

%
%


%

\newpage
\appendix
\section{Additional Results}\label{app:results}
\FloatBarrier
In the following we present additional results of the
feature and time localization capability of the truncated and non-truncated models.
These complement the results shown in Section \ref{sec:results}.
\begin{figure}
  \centering
  \scalebox{0.50}{\input{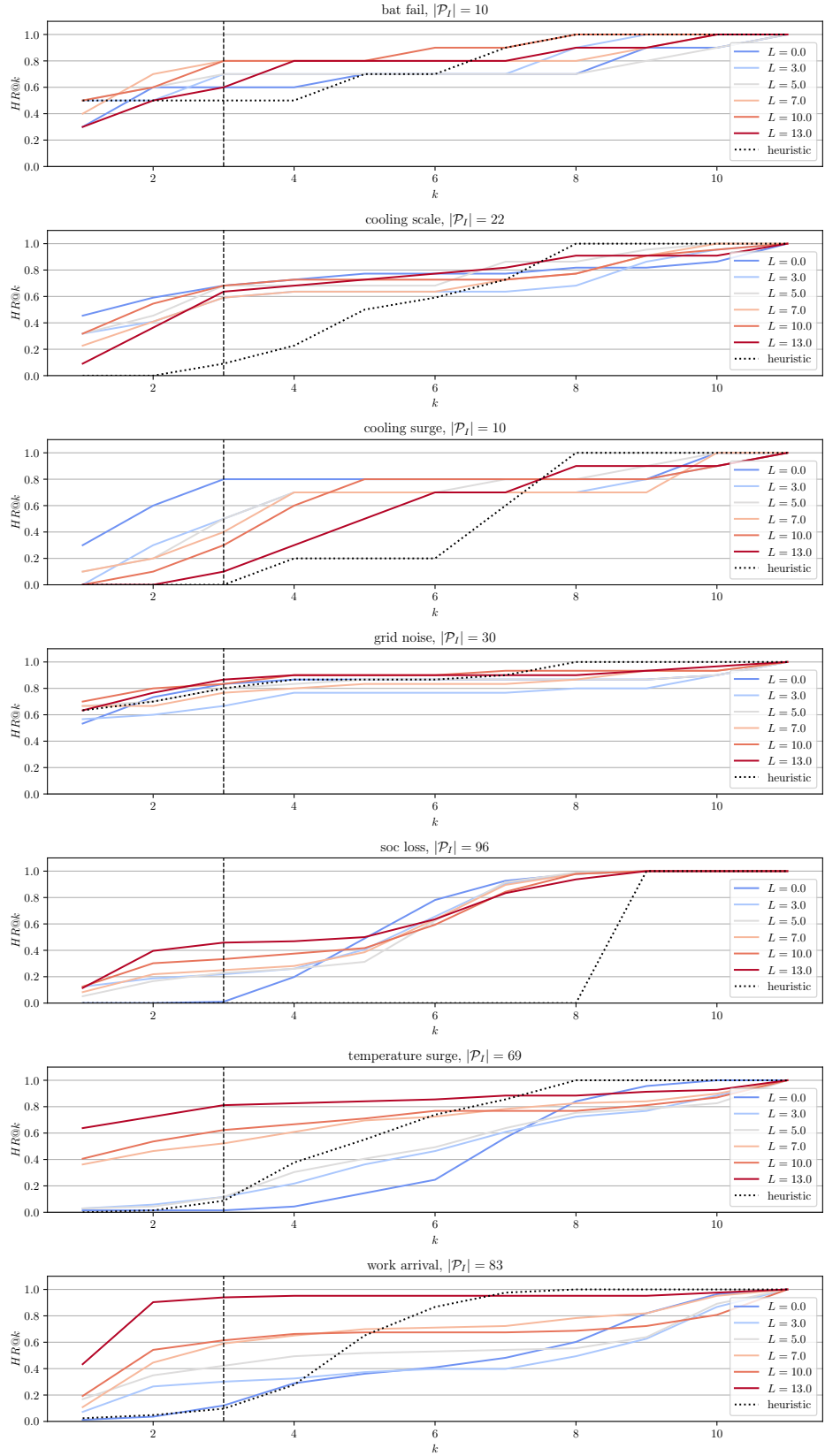}}
  \caption[Hit rate \({\hit}k\) for all injections under \(\Sigma\) aggregation of heuristic and truncated models]{
    Hit rate \({\hit}k\) for all injections under \(\Sigma\) aggregation of heuristic and truncated models.
  }
  \label{fig:result:hitrate:heuristic_vs_trunc_inj_full}
\end{figure}

\begin{figure}
  \centering
  \scalebox{0.50}{\input{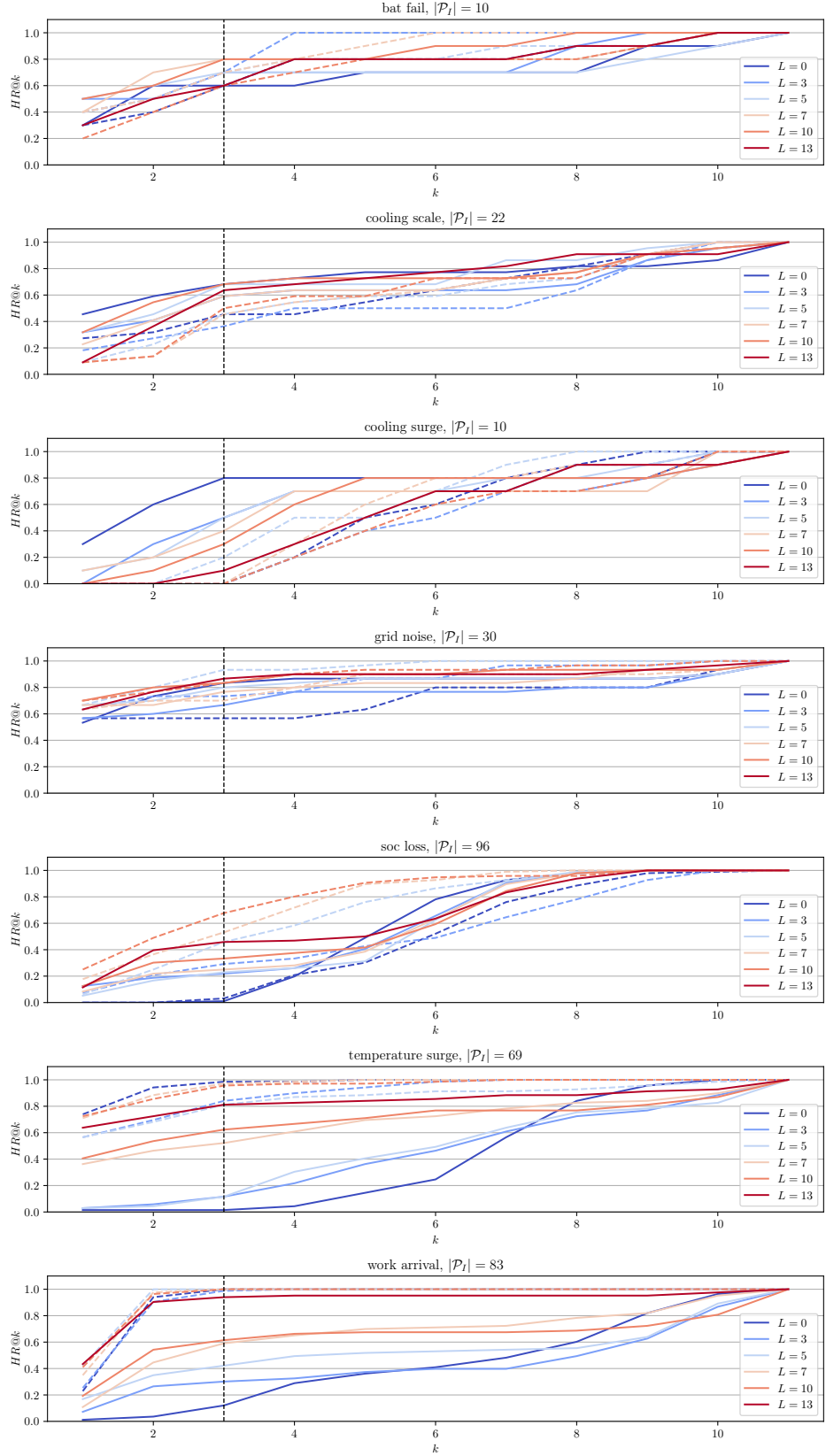}}
  \caption[Hit rate \({\hit}k\) for all injections under \(\Sigma\) aggregation of truncated and non-truncated models]{
    Hit rate \({\hit}k\) for all injections under \(\Sigma\) aggregation of
    truncated (solid lines) and non-truncated models (dashed lines).
  }
  \label{fig:result:hitrate:nontrunc_vs_trunc_inj_full}
\end{figure}

\begin{figure}
  \centering
  \scalebox{0.50}{\input{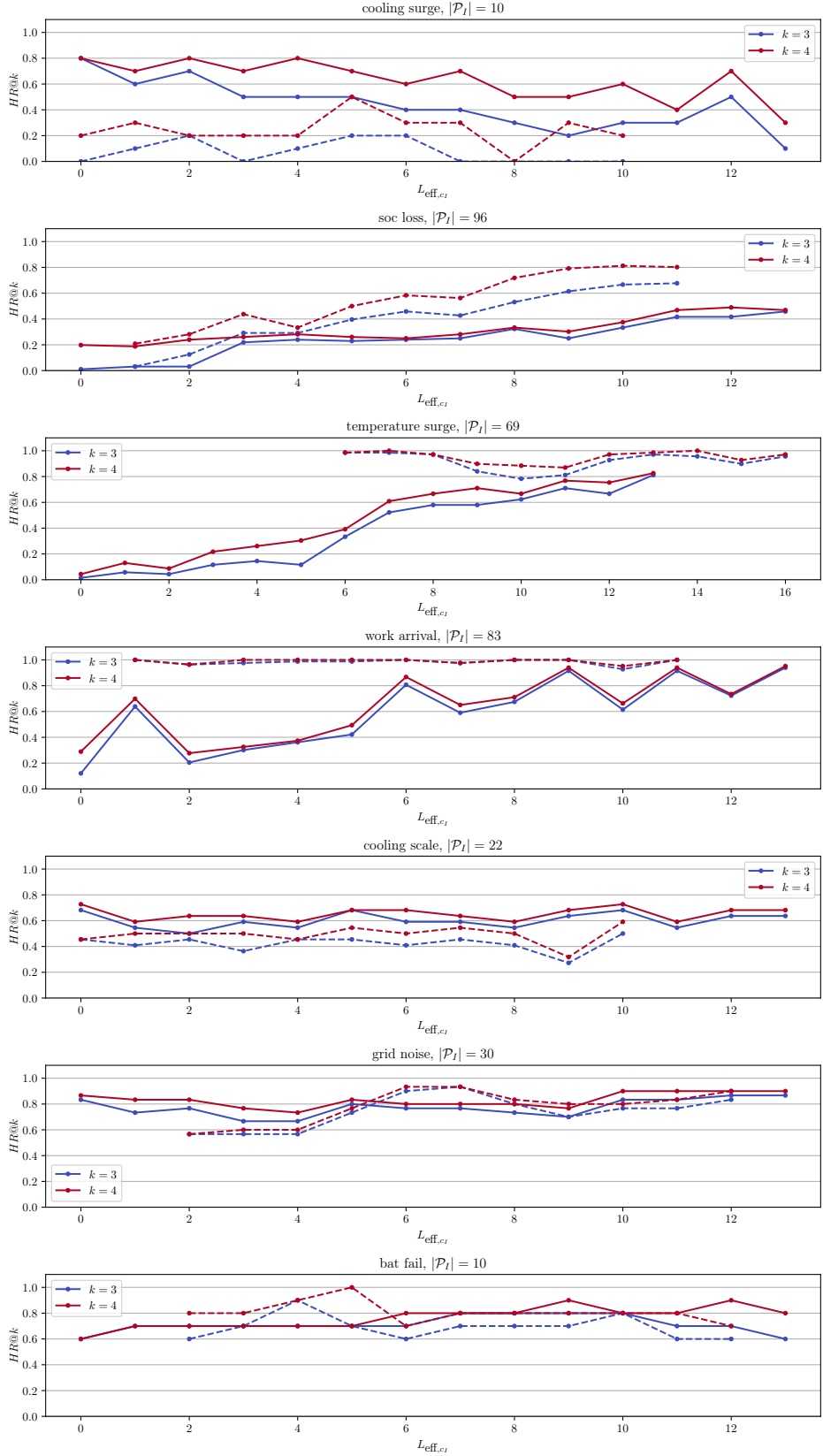}}
  \caption[\({\hit}k\) for all injections in dependence of the effective maximum lag]{
    \({\hit}k\) for all injections under the \(\Sigma\)
    aggregation in dependence of the effective maximum lag \(L_{\eff, c_I}\)
    of the truncated (solid line) and non-truncated (dashed line) models.
  }
  \label{fig:result:hitrate:effectiv_lag_inj_full}
\end{figure}

\begin{figure}
  \centering
  \scalebox{0.45}{\input{figures/bench_time_violin_lag_diff_active_2.pgf}}
  \caption[Box plots of the difference \(d(t_i,\, \hat{t}_i)\) among all peaks caused by immediate injections]{
    Box plots of the difference \(d(t_i,\, \hat{t}_i)\)
    among all peaks caused by immediate injections
    for truncated and non-truncated models.
    The mean difference is given by the dashed blue line.
  }
  \label{fig:result:time_dist:immediate}
\end{figure}

\begin{figure}
  \centering
  \scalebox{0.45}{\input{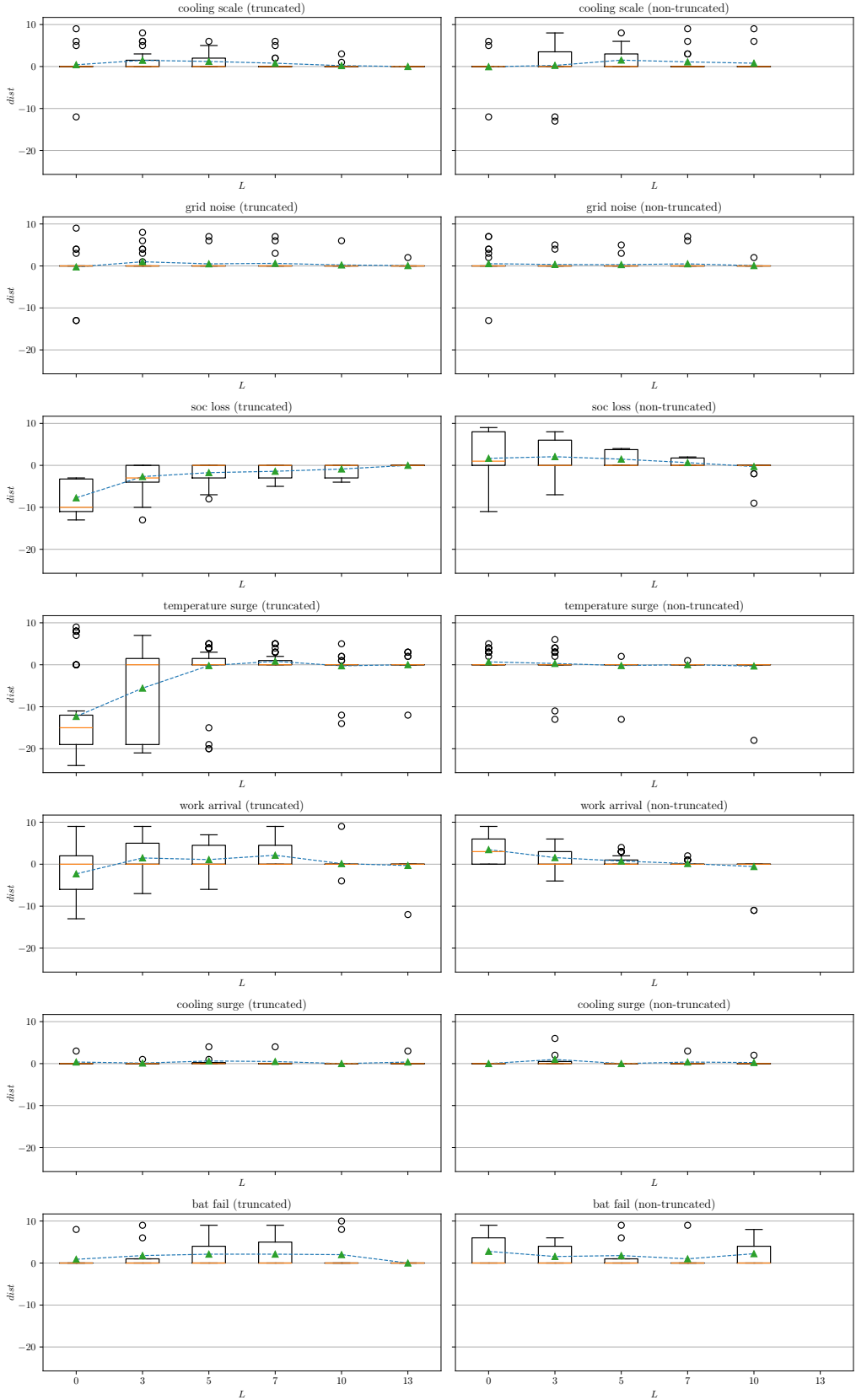}}
  \caption[Box plots of the difference \(d(t_i,\, \hat{t}_i)\) among all peaks with \(\Delta_p \leq 10\)]{
    Box plots of the difference \(d(t_i,\, \hat{t}_i)\)
    among all peaks with \(\Delta_p \leq 10\)
    for truncated and non-truncated models.
    The mean difference is given by the dashed blue line.
  }
  \label{fig:result:time_dist:restricted}
\end{figure}

\FloatBarrier

\section*{Acknowledgements}
This work was funded by the Bavarian Joint Research Program (BayVFP) – Digitization (Funding reference: DIK0294/01).
The research partners AMS-Osram and Economic AI kindly thank the VDI/VDE-IT Munich for the organization and the Free State of Bavaria for the financial support. 

\bibliographystyle{plainnat}
\bibliography{literature.bib}

@book{Shapley1952,
  author="Shapley, Lloyd S.",
  title="A Value for N-Person Games",
  address="Santa Monica, CA",
  year="1952",
  doi="10.7249/P0295",
  publisher="RAND Corporation"
}

@Article{Borges2025,
  author    = {Borges, Dérick G. F. and Coutinho, Eluã R. and Cerqueira-Silva, Thiago and Grave, Malú and Vasconcelos, Adriano O. and Landau, Luiz and Coutinho, Alvaro L. G. A. and Ramos, Pablo Ivan P. and Barral-Netto, Manoel and Pinho, Suani T. R. and Barreto, Marcos E. and Andrade, Roberto F. S.},
  journal   = {BMC Medical Research Methodology},
  title     = {Combining machine learning and dynamic system techniques to early detection of respiratory outbreaks in routinely collected primary healthcare records},
  year      = {2025},
  issn      = {1471-2288},
  month     = apr,
  number    = {1},
  volume    = {25},
  doi       = {10.1186/s12874-025-02542-0},
  publisher = {Springer Science and Business Media LLC},
}

@Article{Guo2024,
  author    = {Guo, Yixiu and Li, Yong and Zhou, Sisi and Zhang, Zhenyu and Wang, Yahui and Xu, Yong and Yang, Xusheng and Li, Zuyi and Shahidehpour, Mohammad},
  journal   = {IEEE Transactions on Smart Grid},
  title     = {Optimal Dispatch for Integrated Energy System Considering Data-Driven Dynamic Energy Hubs and Thermal Dynamics of Pipeline Networks},
  year      = {2024},
  issn      = {1949-3061},
  month     = sep,
  number    = {5},
  pages     = {4537--4549},
  volume    = {15},
  doi       = {10.1109/tsg.2024.3382740},
  publisher = {Institute of Electrical and Electronics Engineers (IEEE)},
}

@Book{Pearl2022,
  author    = {Pearl, Judea},
  publisher = {Cambridge University Press},
  title     = {Causality},
  year      = {2022},
  address   = {Cambridge},
  edition   = {Second edition, reprinted with corrections},
  isbn      = {9780521895606},
  note      = {Hier auch später erschienene, unveränderte Nachdrucke},
  pagetotal = {465},
  ppn_gvk   = {182450313X},
  subtitle  = {Models, reasoning, and inference}
}

@Book{Peters2017,
  author    = {Peters, Jonas},
  editor    = {Dominik Janzing and Bernhard Schölkopf},
  publisher = {The MIT Press},
  title     = {Elements of causal inference},
  year      = {2017},
  address   = {Cambridge, Massachusetts},
  isbn      = {9780262037310},
  pagetotal = {1265},
  ppn_gvk   = {1822189349},
  subtitle  = {Foundations and learning algorithms}
}

@Article{Okati2024,
  author        = {Okati, Nastaran and Mejia, Sergio Hernan Garrido and Orchard, William Roy and Blöbaum, Patrick and Janzing, Dominik},
  title         = {Root Cause Analysis of Outliers with Missing Structural Knowledge},
  year          = {2024},
  month         = jun,
  archiveprefix = {arXiv},
  copyright     = {arXiv.org perpetual, non-exclusive license},
  doi           = {10.48550/ARXIV.2406.05014},
  eprint        = {2406.05014},
  primaryclass  = {stat.ML},
  publisher     = {arXiv}
}

@InProceedings{Budhathoki22a,
  author    = {Budhathoki, Kailash and Minorics, Lenon and Bloebaum, Patrick and Janzing, Dominik},
  booktitle = {Proceedings of the 39th International Conference on Machine Learning},
  title     = {Causal structure-based root cause analysis of outliers},
  year      = {2022},
  editor    = {Chaudhuri, Kamalika and Jegelka, Stefanie and Song, Le and Szepesvari, Csaba and Niu, Gang and Sabato, Sivan},
  month     = {17--23 Jul},
  pages     = {2357--2369},
  publisher = {PMLR},
  series    = {Proceedings of Machine Learning Research},
  volume    = {162},
  url       = {https://proceedings.mlr.press/v162/budhathoki22a.html}
}

@Article{Dawoud2025,
  author        = {Dawoud, Ahmed and Talupula, Shravan},
  title         = {ProRCA: A Causal Python Package for Actionable Root Cause Analysis in Real-world Business Scenarios},
  year          = {2025},
  month         = mar,
  archiveprefix = {arXiv},
  copyright     = {arXiv.org perpetual, non-exclusive license},
  doi           = {10.48550/ARXIV.2503.01475},
  eprint        = {2503.01475},
  primaryclass  = {cs.AI},
  publisher     = {arXiv},
}

@Article{Runge2019,
  author    = {Jakob Runge and Sebastian Bathiany and Erik Bollt and Gustau Camps-Valls and Dim Coumou and Ethan Deyle and Clark Glymour and Marlene Kretschmer and Miguel D. Mahecha and Jordi Mu{\~{n}}oz-Mar{\'{\i}} and Egbert H. van Nes and Jonas Peters and Rick Quax and Markus Reichstein and Marten Scheffer and Bernhard Schölkopf and Peter Spirtes and George Sugihara and Jie Sun and Kun Zhang and Jakob Zscheischler},
  journal   = {Nature Communications},
  title     = {Inferring causation from time series in Earth system sciences},
  year      = {2019},
  month     = {jun},
  number    = {1},
  volume    = {10},
  doi       = {10.1038/s41467-019-10105-3},
  publisher = {Springer Science and Business Media {LLC}}
}

@Article{Runge2019a,
  author    = {Jakob Runge and Peer Nowack and Marlene Kretschmer and Seth Flaxman and Dino Sejdinovic},
  journal   = {Science Advances},
  title     = {Detecting and quantifying causal associations in large nonlinear time series datasets},
  year      = {2019},
  month     = {nov},
  number    = {11},
  volume    = {5},
  doi       = {10.1126/sciadv.aau4996},
  publisher = {American Association for the Advancement of Science ({AAAS})}
}

@Article{Runge2018,
  author    = {Runge, J.},
  journal   = {Chaos: An Interdisciplinary Journal of Nonlinear Science},
  title     = {Causal network reconstruction from time series: From theoretical assumptions to practical estimation},
  year      = {2018},
  issn      = {1089-7682},
  month     = jul,
  number    = {7},
  volume    = {28},
  doi       = {10.1063/1.5025050},
  publisher = {AIP Publishing},
}

@Article{Wang2023,
  author        = {Wang, Dongjie and Chen, Zhengzhang and Ni, Jingchao and Tong, Liang and Wang, Zheng and Fu, Yanjie and Chen, Haifeng},
  title         = {Hierarchical Graph Neural Networks for Causal Discovery and Root Cause Localization},
  year          = {2023},
  month         = feb,
  archiveprefix = {arXiv},
  copyright     = {arXiv.org perpetual, non-exclusive license},
  doi           = {10.48550/ARXIV.2302.01987},
  eprint        = {2302.01987},
  primaryclass  = {cs.LG},
  publisher     = {arXiv}
}

@Article{Liu2021,
  author        = {Liu, Dewei and He, Chuan and Peng, Xin and Lin, Fan and Zhang, Chenxi and Gong, Shengfang and Li, Ziang and Ou, Jiayu and Wu, Zheshun},
  title         = {MicroHECL: High-Efficient Root Cause Localization in Large-Scale Microservice Systems},
  year          = {2021},
  month         = may,
  pages         = {338--347},
  archiveprefix = {arXiv},
  booktitle     = {2021 IEEE/ACM 43rd International Conference on Software Engineering: Software Engineering in Practice (ICSE-SEIP)},
  copyright     = {Creative Commons Attribution Non Commercial No Derivatives 4.0 International},
  date          = {2021-03-01},
  doi           = {10.1109/icse-seip52600.2021.00043},
  eprint        = {2103.01782},
  primaryclass  = {cs.SE},
  publisher     = {IEEE}
}

@Article{Lin2024,
  author        = {Lin, Cheng-Ming and Chang, Ching and Wang, Wei-Yao and Wang, Kuang-Da and Peng, Wen-Chih},
  title         = {Root Cause Analysis In Microservice Using Neural Granger Causal Discovery},
  year          = {2024},
  month         = feb,
  archiveprefix = {arXiv},
  copyright     = {Creative Commons Attribution Share Alike 4.0 International},
  doi           = {10.48550/ARXIV.2402.01140},
  eprint        = {2402.01140},
  primaryclass  = {cs.LG},
  publisher     = {arXiv}
}

@Article{Ikram2022,
  author    = {Ikram, Azam and Chakraborty, Sarthak and Mitra, Subrata and Saini, Shiv and Bagchi, Saurabh and Kocaoglu, Murat},
  booktitle = {Advances in Neural Information Processing Systems},
  title     = {Root Cause Analysis of Failures in Microservices through Causal Discovery},
  year      = {2022},
  editor    = {S. Koyejo and S. Mohamed and A. Agarwal and D. Belgrave and K. Cho and A. Oh},
  pages     = {31158--31170},
  publisher = {Curran Associates, Inc.},
  volume    = {35},
  url       = {https://proceedings.neurips.cc/paper_files/paper/2022/file/c9fcd02e6445c7dfbad6986abee53d0d-Paper-Conference.pdf}
}

@Article{Strobl2024,
  author    = {Strobl, Eric V.},
  journal   = {Journal of Biomedical Informatics},
  title     = {Counterfactual formulation of patient-specific root causes of disease},
  year      = {2024},
  issn      = {1532-0464},
  month     = feb,
  pages     = {104585},
  volume    = {150},
  doi       = {10.1016/j.jbi.2024.104585},
  publisher = {Elsevier BV},
}

@Article{Bloebaum2024,
  author  = {Patrick Bl{{\"o}}baum and Peter G{{\"o}}tz and Kailash Budhathoki and Atalanti A. Mastakouri and Dominik Janzing},
  journal = {Journal of Machine Learning Research},
  title   = {DoWhy-GCM: An Extension of DoWhy for Causal Inference in Graphical Causal Models},
  year    = {2024},
  number  = {147},
  pages   = {1--7},
  volume  = {25},
  url     = {http://jmlr.org/papers/v25/22-1258.html}
}

@Article{Sharma2020,
  author        = {Sharma, Amit and Kiciman, Emre},
  title         = {DoWhy: An End-to-End Library for Causal Inference},
  year          = {2020},
  month         = nov,
  archiveprefix = {arXiv},
  copyright     = {arXiv.org perpetual, non-exclusive license},
  doi           = {10.48550/ARXIV.2011.04216},
  eprint        = {2011.04216},
  primaryclass  = {stat.ME},
  publisher     = {arXiv}
}
\end{document}